%% file: main_fixed.tex
\pdfoutput=1

\documentclass[11pt]{article}
\usepackage[table]{xcolor} 

\usepackage[]{acl}

\usepackage{times}
\usepackage{latexsym}
\usepackage[T1]{fontenc}

\usepackage[utf8]{inputenc}
\usepackage{enumitem}

\usepackage{amssymb}
\usepackage{subcaption}

\usepackage{pgfplots}
\usepackage{xcolor}
\usepackage{tikz}
\usepackage{bchart}

\usepackage{placeins} 

\usepackage{tipa}
\DeclareUnicodeCharacter{01C1}{\textipa{||}}

\newcommand{\dsbox}[1]{\textcolor{#1}{\rule{1.5ex}{1.5ex}}}

\usetikzlibrary{shapes.geometric}
\definecolor{ptcolor}{HTML}{FFD700}  
\definecolor{ftcolor}{HTML}{17BECF}  

\definecolor{olivegreen}{RGB}{107,142,35}
\definecolor{lightolivegreen}{RGB}{157,192,105}
\definecolor{customgreen}{rgb}{0.13, 0.55, 0.13}

\usepackage{booktabs}
\usepackage{threeparttable}

\definecolor{findOptimalPartition}{HTML}{D7191C}
\definecolor{storeClusterComponent}{HTML}{FDAE61}
\definecolor{dbscan}{HTML}{ABDDA4}
\definecolor{constructCluster}{HTML}{2B83BA}

\usepackage{microtype}
\usepackage{graphicx}
\usepackage{booktabs} 
\usepackage{tabularx}
\usepackage{multirow} 
\usepackage{hyperref}
\usepackage{lipsum}
\usepackage{xltabular}
\usepackage{longtable}
\usepackage{cals}
\usepackage{xcolor,colortbl}
\usepackage[normalem]{ulem}
\usepackage{pifont}

\usepackage{soul}
\usepackage{multirow}
\usepackage{amsmath}
\usepackage{amssymb}
\usepackage{listings}
\usepackage{algorithm}
\usepackage{algpseudocode}
\usepackage{amssymb}
\usepackage{cuted}

\usepackage{appendix}
\usepackage{array} 
\newcolumntype{H}{>{\setbox0=\hbox\bgroup}c<{\egroup}@{}}

\usepackage[normalem]{ulem}  
\usepackage{soul}            

\newcommand{\linecolor}[2]{\setulcolor{#1}\ul{#2}}

\usepackage{hyperref} 

\definecolor{speech}{HTML}{70D6FF}
\definecolor{government}{HTML}{FF70A6}
\definecolor{benchmarkks}{HTML}{FF9770}
\definecolor{stories}{HTML}{FFD670}
\definecolor{news}{HTML}{E9FF70}
\definecolor{health}{HTML}{a5ffd6}
\definecolor{wikipedia}{HTML}{ff69eb}
\definecolor{religious}{HTML}{7371fc}
\definecolor{web}{HTML}{D62728}

\definecolor{glotlid}{HTML}{ff686b}
\definecolor{afrolid}{HTML}{9467BD}
\definecolor{simbatext}{HTML}{C5B0D5}
\definecolor{fineweb}{HTML}{ffe45e}
\definecolor{flores}{HTML}{8C564B}
\definecolor{mafand}{HTML}{C49C94}
\definecolor{smol}{HTML}{E377C2}
\definecolor{mcs}{HTML}{F7B6D2}
\definecolor{openlid}{HTML}{7F7F7F}
\definecolor{udhr}{HTML}{C7C7C7}

\usepackage{xspace}

\def \numlanguages{$640$}

\usepackage{pgfplots}
\pgfplotsset{compat=1.18}
\usepackage{arydshln}

\usepackage{makecell}
\usepackage{etoolbox}
\usepackage{hyperref}

\newcommand{\ourframework}{\textsc{AfroScope}}
\newcommand{\ourdata}{\textsc{AfroScope-Data}}
\newcommand{\ourmodel}{\textsc{AfroScope-Models}}
\newcommand{\ourmirror}{\textsc{AfroScope-Mirror}}

\title{\raisebox{-0.29\height}{\includegraphics[scale=0.025]{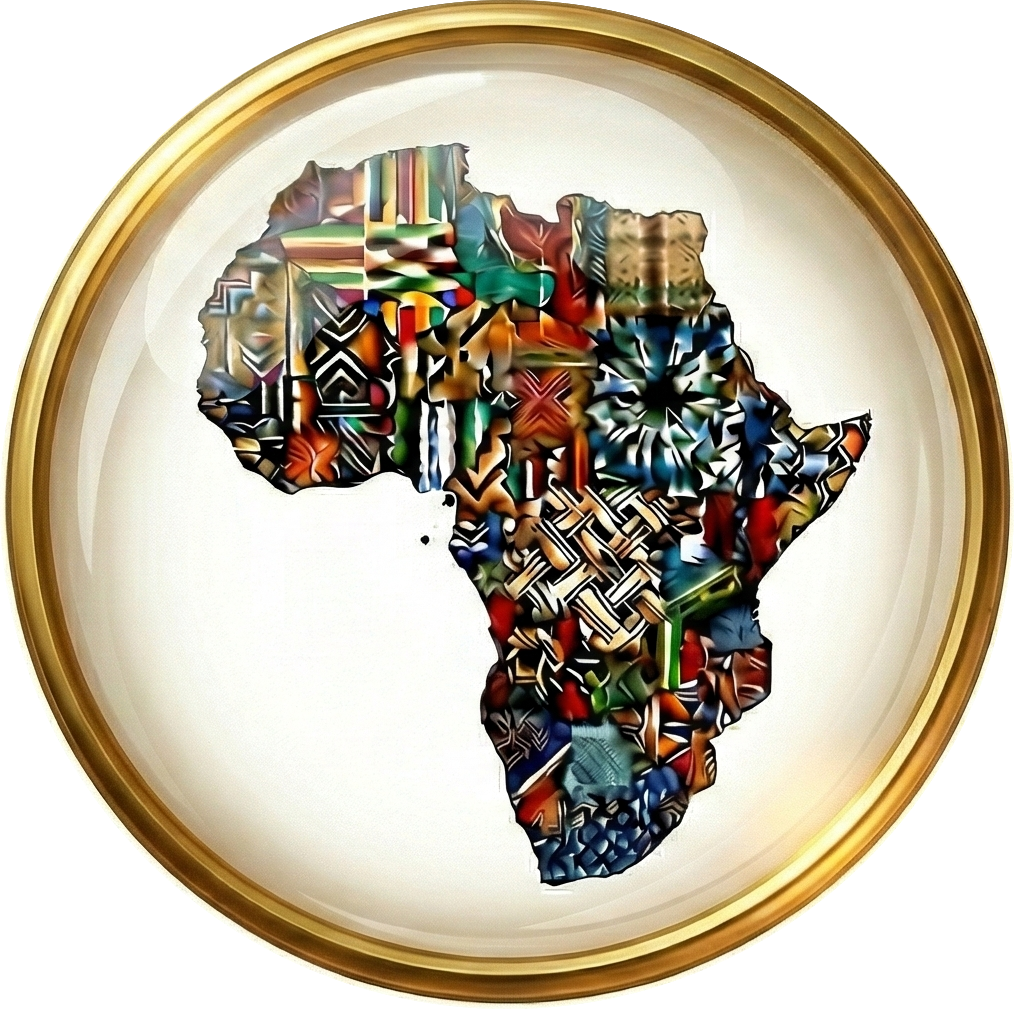}}~\ourframework: A Framework for Studying the Linguistic Landscape of Africa\xspace}

\author{\normalsize Sang Yun Kwon$^{\xi}$ ~~ AbdelRahim Elmadany$^{\xi}$  ~~ Muhammad Abdul-Mageed$^{\xi,\lambda}$\\
\normalsize $^{\xi}$The University of British Columbia ~~ $^{\lambda}$Canada Research Chair in NLP and ML \\%
  \texttt{\normalsize \{skwon01@mail.,a.elmadany@,muhammad.mageed@\}ubc.ca}}

\begin{document}
\maketitle
\begin{strip}
  \centering
  \includegraphics[width=\textwidth]{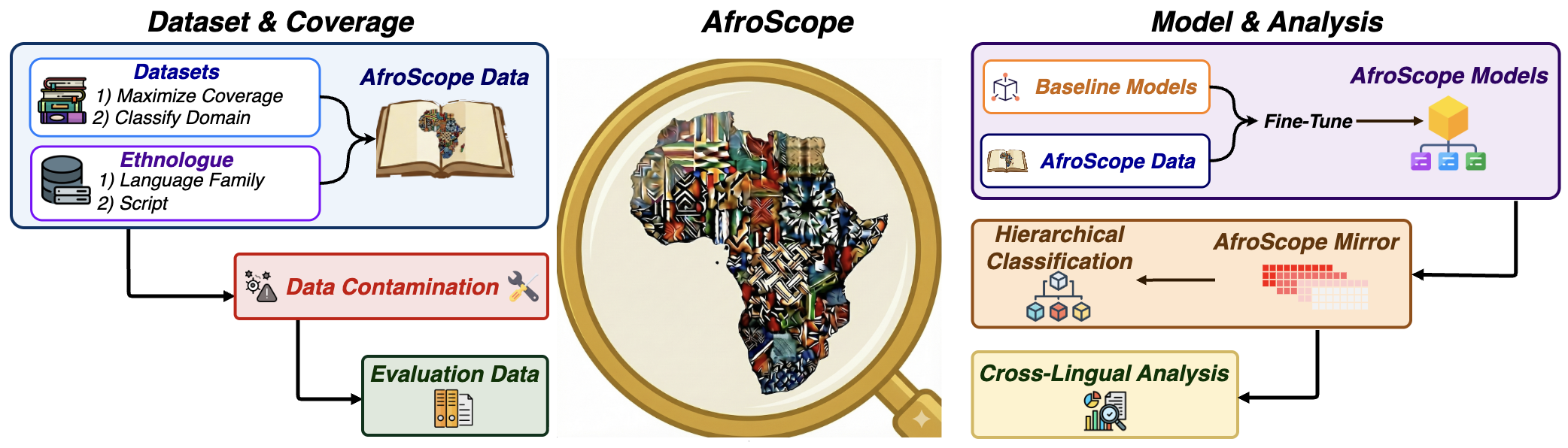}
  \vspace{-1.5em}
  \captionof{figure}{Overview of the \textsc{AfroScope} framework. 
\textbf{Dataset \& Coverage} constructs \textsc{AfroScope-Data} by aggregating multilingual datasets and metadata on language families, scripts, and domains, while applying contamination analysis to support clean evaluation. \textbf{Model \& Analysis} fine-tunes baseline models to produce \textsc{AfroScope-Models}, introduces \textsc{AfroScope-Mirror} for hierarchical disambiguation of closely related languages, and studies cross-lingual transfer and robustness across African languages.}
  \label{fig:main_fig}
\end{strip}

\input{sections/abstract}

\input{sections/introduction}

\input{sections/literature_review}
\input{sections/data_collection}
\input{sections/experimental_setup}

\input{sections/eval_results}

\input{sections/discussions}

\input{sections/conclusion}
\input{sections/limit_ethics}

\normalem
\bibliography{custom}
\input{sections/appendix}

\end{document}

%% file: sections/abstract.tex
\begin{abstract}

Language Identification (LID), the task of determining the language of a given text, is a fundamental preprocessing step that shapes the reliability of downstream NLP applications. While recent work has expanded African LID, existing systems remain limited in both language coverage and fine-grained discrimination among closely related languages and varieties. We introduce \ourframework, a unified framework for African LID that includes \ourdata, a dataset covering \numlanguages~languages, and \ourmodel, a suite of strong LID models with broad African language coverage. To address persistent confusions among closely related languages, we propose a hierarchical classification approach that leverages \ourmirror, a specialized embedding model for targeted disambiguation. This approach improves macro-F\textsubscript{1} by $1.57$ points on the confusable subset compared to our best base model. We further analyze cross-lingual transfer and domain effects, showing how language-family structure, script compatibility, and domain coverage shape LID performance. We position African LID as an enabling technology for large-scale measurement of Africa's linguistic landscape in digital text and release \ourdata\ and \ourmodel\ online.\footnote{\href{https://github.com/UBC-NLP/AfroScope}{UBC-NLP/AfroScope}}


\end{abstract}

%% file: sections/introduction.tex
\section{Introduction}\label{sec:introduction}

Language Identification (LID), the task of determining the language of a given text, is a foundational step in curating multilingual corpora from web crawls~\cite{penedo2025fineweb2,foroutan2025conlid}. LID errors propagate to downstream stages such as tokenization~\cite{duvenhage2017improved}, filtering~\cite{grattafiori2024llama, li2024datacomp}, and data scheduling for multilingual pretraining~\cite{conneau-etal-2020-unsupervised, de-gibert-etal-2024-new, laurenccon2022bigscience}. Crucially, LID systems determine not only how reliably each language is predicted but also the \textit{scope} of identifiable languages. 
If a language is out of scope, its text is either dropped or misattributed to an in-scope language, distorting corpus composition and downstream evaluation~\cite{costa2022no,adebara2022afrolid}.

Although LID is often treated as largely solved~\cite{mcnamee2005language, blodgett2017racial}, recent evaluations show that performance remains uneven across languages and domains~\cite{ojo2025divers, suarez2026commonlid}. LID for African languages remains especially challenging. Existing benchmarks cover only a small fraction of the continent's languages, and even within this limited coverage, performance often lags behind that of higher-resource languages. At the same time, major web-crawled corpora for African languages suffer from systematic quality issues~\cite{kreutzer2022quality}, including noisy or unusable text, misattributed documents~\cite{alabi-etal-2020-massive}, and heavy concentration in religious or translated material that does not adequately reflect everyday language use~\cite{kargaran-etal-2023-glotlid}.
These artifacts degrade downstream performance and inflate apparent coverage—a form of \textit{representation washing}~\cite{burchell-etal-2023-open} that reinforces disparities in language technology~\cite{blasi-etal-2022-systematic}.
Recent LID systems~\cite{kargaran-etal-2023-glotlid, foroutan2025conlid}, including African-focused models~\cite{adebara2022afrolid}, have made important progress. However, two central gaps remain: \textit{scope}, i.e., the set of African languages that can be reliably identified, and \textit{granularity}, i.e., the ability to distinguish closely related languages and varieties. We address these gaps with \ourframework{}, a unified framework for African LID. As shown in Figure~\ref{fig:main_fig}, \ourframework{} consists of three main contributions:

\textbf{(i) Coverage-oriented data and models.} We introduce \ourdata{}, a large-scale multilingual dataset curated to expand both language and domain coverage for African LID. \ourdata{} spans \numlanguages~languages across multiple orthographies and domains (\S\ref{sec:afroscope}), enabling a more comprehensive assessment of African LID performance, including fine-grained analysis across domains. Using \ourdata{}, we train \ourmodel{}, a family of LID models that outperform prior African LID baselines across our evaluation setting (\S\ref{sec:experimental_setup}).

\textbf{(ii) Hierarchical disambiguation of closely related languages.} We show that genetically related and geographically proximate languages are a major source of false positives, especially when labels correspond to closely related varieties or macrolanguage members. To address this, we introduce \ourmirror{}, a lightweight contrastive embedding model specialized for frequently confused language groups, and use it in a hierarchical inference procedure that performs targeted disambiguation only when fine-grained separation is needed  (\S\ref{sec:hierarchical}). This design improves discrimination among closely related languages while preserving broad-coverage LID.

\textbf{(iii) Transfer and robustness analysis.} Leveraging \ourdata{}, we analyze LID performance by coverage and domain (\S\ref{sec:result}), and study multilingual transfer effects---including positive transfer and negative interference---as a function of language family structure and script overlap (\S\ref{sec:transfer}). These analyses provide practical guidance for building, evaluating, and curating robust African LID systems.


%% file: sections/literature_review.tex
\begin{figure*}[!htb]
\centering
\minipage{0.33\textwidth}
  \includegraphics[width=\linewidth]{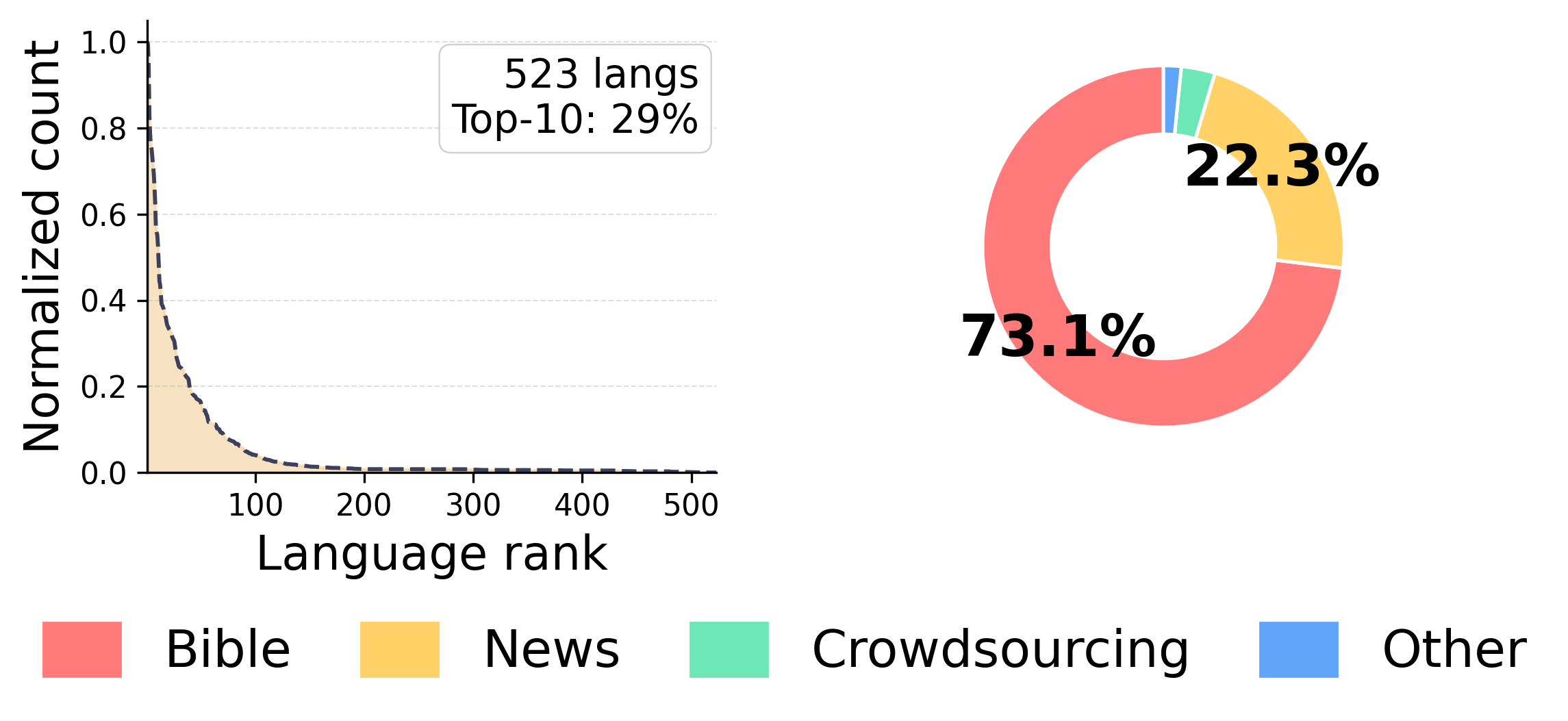}
  \subcaption{GlotLID-C dataset}\label{fig:glotlid_dist}
\endminipage\hfill
\minipage{0.33\textwidth}
  \includegraphics[width=\linewidth]{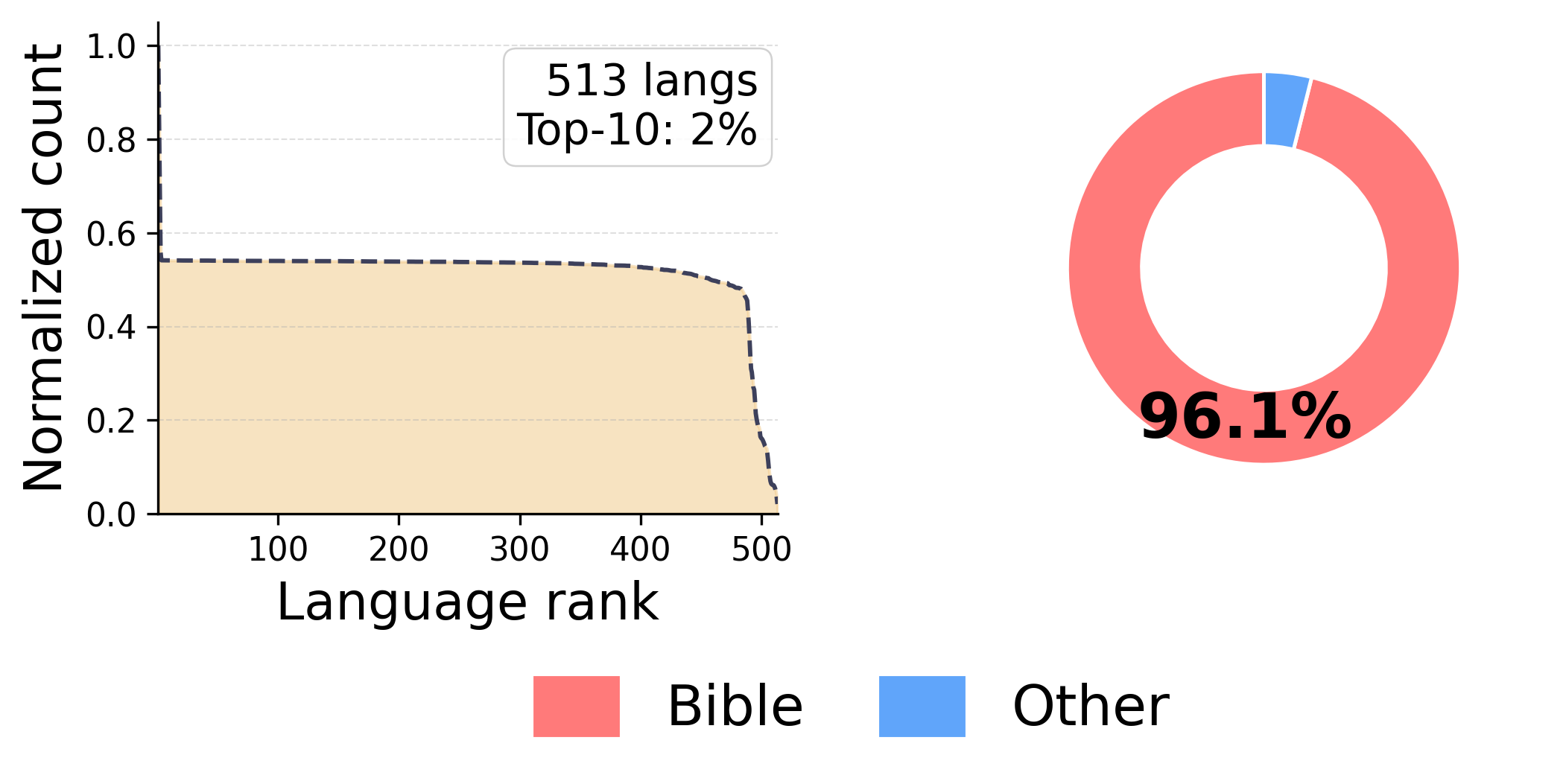}
  \subcaption{AfroLID dataset}\label{fig:afrolid_dist}
\endminipage\hfill
\minipage{0.33\textwidth}
  \includegraphics[width=\linewidth]{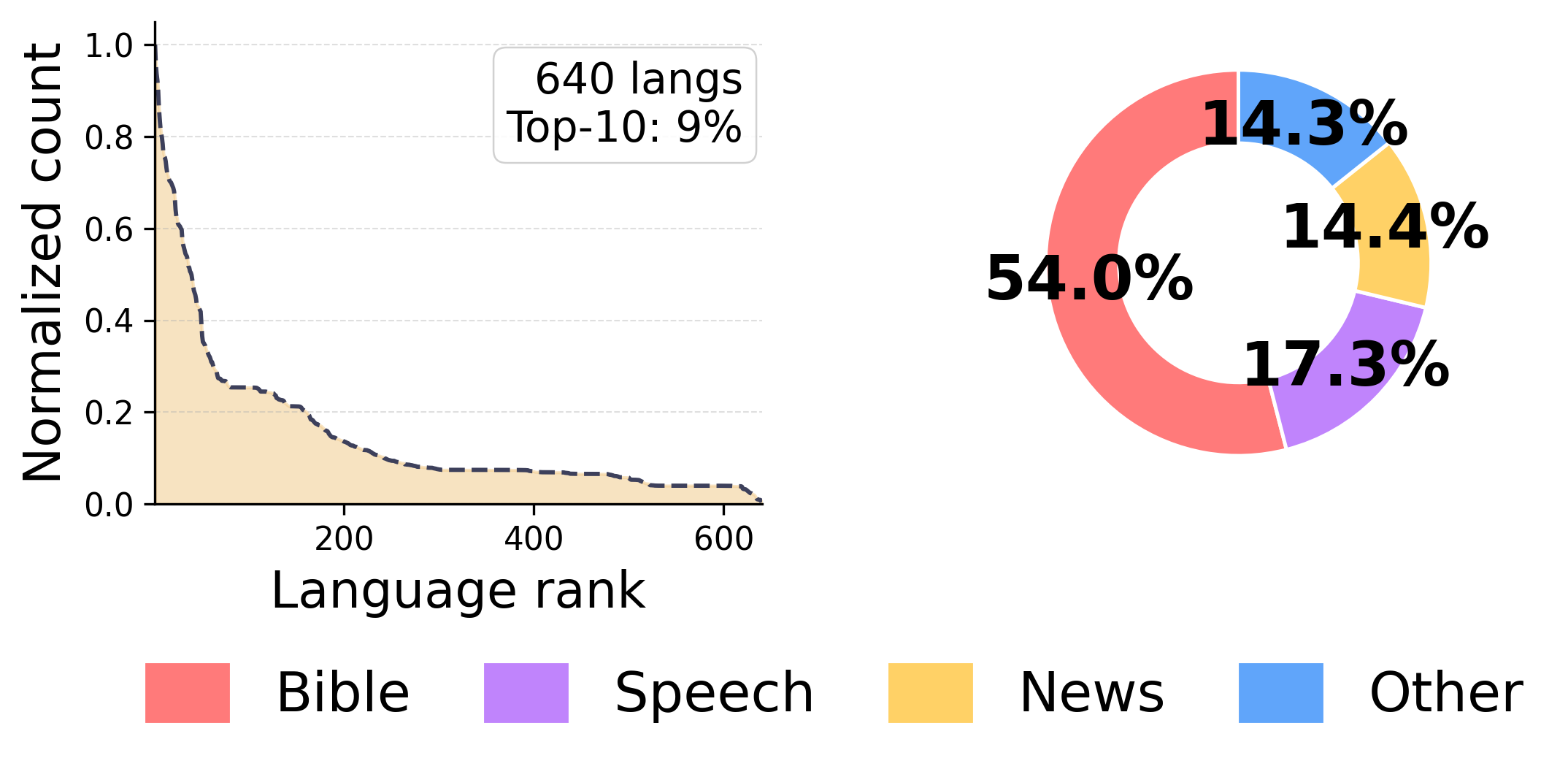}
  \subcaption{\ourdata}\label{fig:afroscope_dist}
\endminipage
\caption{Language-resource and domain distributions for GlotLID-C, AfroLID, and \textsc{AfroScope-Data}. Each pair of plots shows the language distribution of sentence counts across languages (\textit{left}) and the corresponding domain composition (\textit{right}). \textsc{AfroScope-Data} reduces language imbalance and domain concentration relative to the source corpora. Domains accounting for less than $3$\% of a dataset are grouped under \textit{Other}.}
\label{longtail_domain}
\end{figure*}

%


\section{Related Works}\label{sec:lit_review} 

\paragraph{Linguistic diversity in Africa.}
Africa is among the most linguistically diverse regions, spanning many language families and typological profiles~\citep{HeineNurse2000,eberhard2021ethnologue}. For NLP systems, this diversity manifests in phenomena that directly stress corpus curation and LID, including rich morphology, orthographic variation, and pervasive multilingual practices such as code-switching~\cite{ojo2025divers, hussen2025state}. In addition, language varieties with fluid boundaries complicate labeling and evaluation~\citep{abdulmumin-etal-2024-correcting, alabi2025charting}, particularly for diglossic African languages like Arabic. Recent work has responded with new resources, benchmarks, and African-focused models, which we discuss---along with related challenges---in Appendix~\ref{appdx_sec:lit_review_diversity}.


\paragraph{Data authenticity and corpus quality.}
Large multilingual corpora frequently contain misattributed text, ambiguous language codes, and other quality issues that disproportionately affect low-resource settings~\citep{kreutzer2022quality}. Prior studies of widely used multilingual resources and pipelines document systematic noise and labeling errors~\cite{banon2020paracrawl,schwenk2021wikimatrix,xue2021mt5}, and emphasize the role of LID quality and preprocessing in mitigating such artifacts~\citep{agarwal2023limit}. Improving authenticity is therefore central to building reliable and culturally representative language technologies~\citep{ojo2023good,zhong2024opportunities,alhanai2025bridging}, motivating dataset construction that explicitly controls for coverage, domain diversity, and contamination.

\paragraph{Progress in African language identification.}
Recent evaluations show that LID coverage for African languages remains narrow and performance lags substantially~\cite{suarez2026commonlid}. Despite progress from FastText-based systems~\cite{joulin2016bag, kargaran-etal-2023-glotlid, burchell-etal-2023-open} to African-focused transformer-based models~\cite{adebara2022afrolid, adebara2022serengeti, adebara2024cheetah}, as well as methodological advances through contrastive learning~\cite{foroutan2025conlid} and hierarchical approaches~\cite{agarwal2023limit}, targeted efforts for African LID remain necessary to address problems specific to African languages, such as domain variation and closely related varieties~\cite{salamanca2026tiny}.

%% file: sections/data_collection.tex
\input{tables/afroscope_stats_old}

\section{Addressing Coverage and Domain Skew:  \textit{\ourdata}}\label{sec:afroscope}
Building robust LID systems for African languages requires data that is broad across two axes: \textit{language coverage} and \textit{domain coverage}. Existing resources cover only a small subset of African languages, leaving many underrepresented or entirely absent and making it difficult to measure true progress. Even among covered languages, sentence counts are highly concentrated in a small number of relatively high-resource languages (Figure~\ref{fig:glotlid_dist}). Domain coverage is similarly skewed: religious texts, predominantly Bible translations, account for $73.10$\% and $94.70$\% of sentences in GlotLID (Figure~\ref{fig:glotlid_dist}) and AfroLID (Figure~\ref{fig:afrolid_dist}), respectively. This narrow distribution limits generalization to heterogeneous digital text, including web text, administrative documents, and transcribed speech. 

To address these challenges, we curate \ourdata~(Figure~\ref{fig:afroscope_dist}) using a strategy guided by two objectives: \textit{(i) maximizing language coverage} to reduce \textit{out-of-model
cousin} errors~\cite{caswell-etal-2020-language,kreutzer2022quality}, where text from an unsupported language is misattributed to the closest supported relative; and \textit{(ii) increasing domain diversity} to mitigate the narrow domain concentration in available African language data.

\ourdata{} (Table~\ref{tab:dataset_stats}) spans \numlanguages{} African languages across eight language families, seven scripts, and eight domains. To the best of our knowledge, \ourdata{} provides the broadest publicly described coverage for African LID in terms of the joint number of African language labels and domain categories.

\subsection{Data Curation and Sampling}\label{subsec:data_curation}
We compile \ourdata{} from publicly described sources, prioritizing datasets that provide sufficient provenance metadata for assigning coarse domain labels. To maximize coverage, we use \texttt{GlotLID-C}~\cite{kargaran-etal-2023-glotlid}, \texttt{AfroLID}~\cite{adebara2022afrolid}, and \texttt{SimbaText}~\cite{elmadany2025voice} as training sources, given their breadth and metadata availability (Table~\ref{tab:dataset_stats}). 


To mitigate language imbalance, we sample a fixed-size training corpus using temperature smoothing, following prior multilingual practice~\cite{arivazhagan2019massively, costa2022no, burchell-etal-2023-open}. For a language $l$ with corpus fraction $p_l$, we sample proportional to $p_l^{\alpha}$, where $\alpha=0.3$. We apply the same smoothing hierarchically at the domain level: within each language's budget, a domain $d$ with intra-language fraction $p_d$ is sampled proportional to $p_d^{\alpha}$. This reduces the dominance of high-frequency domains, such as religious texts, and increases the representation of lower-frequency ones.

Through this sampling strategy, \ourdata{} mitigates imbalance along both axes (Figure~\ref{fig:afroscope_dist}). The top-10 languages account for only $9$\% of sentences, down from $29$\% in GlotLID-C. Bible-derived content is reduced to $53.5$\%, compared to $73.1$\% and $94.7$\% in GlotLID-C and AfroLID, respectively. We provide more details regarding data curation and sampling in Appendix~\ref{app_sec:data_curation}.

\subsection{Family, Script, and Domain Coverage}\label{subsec:data_domains} 


\paragraph{Language family.} \ourdata{} spans eight high-level groupings: \textit{Afro-Asiatic, Austronesian, Creole, Indo-European, Khoe-Kwadi, Mixed language, Niger-Congo,} and \textit{Nilo-Saharan}. This diversity reduces reliance on cues from dominant families such as \textit{Niger-Congo} and supports evaluation of cross-family generalization. We use this hierarchy in our transfer analyses (\S\ref{sec:transfer}).

\input{tables/better_eval}

\paragraph{Script.} We include seven writing systems: \textit{Latin} (\texttt{Latn}), \textit{Arabic} (\texttt{Arab}), \textit{Ge'ez} (\texttt{Ethi}), \textit{N'Ko} (\texttt{Nkoo}), \textit{Tifinagh} (\texttt{Tfng}), \textit{Coptic} (\texttt{Copt}), and \textit{Vai} (\texttt{Vaii}). We explicitly label scripts for each language (\textit{language\_script}), as individual languages may employ multiple writing systems (e.g., \textit{gof, ttq}), enabling script-aware evaluation and analysis of orthographic variation. 

\paragraph{Domain.} 

To analyze domain effects, we first adopt the source-domain categorization from GlotLID~\cite{kargaran-etal-2023-glotlid}, which groups sources into coarse categories such as bible, news, web, government, and social text. We extend this mapping to sources absent from the original GlotLID taxonomy, using the origin metadata each dataset provides. This yields a compact set of domains: \textit{Speech, Human Rights, Crowdsourcing, News, Bible, Web, Government}, and \textit{Other}. We assign each sentence a domain label through this source-level mapping and use the labels for controlled evaluation by domain; we discuss domain effects in detail in \S\ref{sec:result}.

%% file: tables/afroscope_stats_old.tex
\begin{table*}[!ht]
\centering
\resizebox{0.80\textwidth}{!}{%

\begin{tabular}{clllcccc}
\toprule
& \multicolumn{2}{l}{\textbf{Dataset}} & \textbf{Sent.} & \textbf{Lang.} & \textbf{Family} & \textbf{Script} & \multicolumn{1}{c}{\textbf{Domain    }} \\
\midrule

\multirow{5}{*}{\rotatebox{90}{\textsc{Train}}}
& GlotLID-C~\cite{kargaran-etal-2023-glotlid}   &  & $60{,}682{,}541$        & $523$    & $7$ & $5$ & \dsbox{religious} \dsbox{web} \dsbox{benchmarkks} \dsbox{news} \dsbox{stories} \dsbox{wikipedia} \\
& AfroLID~\cite{adebara2022afrolid}           &  & $1{,}682{,}541$        & $513$  & $7$ & $5$ & \dsbox{religious} \dsbox{web} \dsbox{stories} \dsbox{news} \\
& SimbaText~\cite{elmadany2025voice}          &  & $382{,}541$        & $101$   & $5$ & $4$ & \dsbox{speech} \dsbox{news} \dsbox{web} \\ \cmidrule{2-8}
& \multirow{2}{*}{\textbf{\ourdata}}
  & \textit{Train} & $5{,}952{,}575$ & \multirow{2}{*}{$640$}
                       & \multirow{2}{*}{$8$}
                       & \multirow{2}{*}{$7$}
                       & \multirow{2}{*}{ \dsbox{speech} \dsbox{religious} \dsbox{web} \dsbox{government} \dsbox{benchmarkks} \dsbox{stories} \dsbox{wikipedia} \dsbox{health}} \\
&  & \textit{Dev}   & $463{,}875$     &                    &                    &                    &                     \\
\midrule

\multirow{7}{*}{\rotatebox{90}{\textsc{Evaluation}}}
&  FLORES$+$~\cite{nllb-24}                    &    & $108{,}486$  & $54$    & $5$ & $4$ & \dsbox{benchmarkks} \\
& MAFAND~\cite{adelani-etal-2022-thousand}   &   & $54{,}795$   & $21$    & $4$ & $2$ & \dsbox{news} \\
& SmolSent~\cite{caswell2025smol}               &    & $10{,}872$   & $53$    & $6$ & $4$ & \dsbox{benchmarkks} \\
& MCS-350~\cite{agarwal2023limit}            &       & $94{,}894$   & $141$    & $6$ & $3$ & \dsbox{health} \\
& UDHR~\cite{kargaran-etal-2023-glotlid}      &  &  $6{,}696$    & $117$   & $6$ & $4$ & \dsbox{government} \\
& CommonLID~\cite{suarez2026commonlid}      &  &  $10{,}696$    & $25$   & $3$ & $3$ &  \dsbox{web} \\ \cmidrule{2-8}
& \textbf{\ourdata} & \textit{Test}  & $232{,}563$     & $640$ & $8$ & $7$ & \dsbox{speech} \dsbox{religious} \dsbox{web} \dsbox{government} \dsbox{benchmarkks} \dsbox{stories} \dsbox{wikipedia} \dsbox{health} \\
\bottomrule

\multicolumn{8}{c}{ \dsbox{speech} Speech \quad \dsbox{government} Human Rights \quad \dsbox{benchmarkks} Crowdsourcing \quad  \dsbox{news} News \quad  \dsbox{religious} Bible \quad \dsbox{web} Web } \dsbox{wikipedia} Government \quad \quad \dsbox{health} Other \quad  \\
\end{tabular}%
 }


\vspace{1mm} 
\caption{Summary statistics for the datasets used to construct and evaluate \textsc{AfroScope-Data}. We report the number of sentences (\textit{Sent.}), language labels (\textit{Lang.}), high-level language-family groupings (\textit{Family}), scripts (\textit{Script}), and domains (\textit{Domain}). Colored squares show the domain categories represented in each source. Details on source composition and domain mapping are provided in Appendix~\ref{app_sec:domain_meta}.}
\captionsetup{labelformat=empty,font=footnotesize,justification=raggedright}

%


\label{tab:dataset_stats}
\end{table*}

%% file: tables/better_eval.tex
\begin{table*}[t]
\centering
\resizebox{\textwidth}{!}{%
\begin{tabular}{l cc cc cc cc cc cc cc}
\toprule
\multirow{3}{*}{\centering\textbf{Models}}
& \multicolumn{2}{c}{\textbf{FLORES+}} 
& \multicolumn{2}{c}{\textbf{UDHR (Human Rights)}} 
& \multicolumn{2}{c}{\textbf{CommonLID (Web)}} 
& \multicolumn{2}{c}{\textbf{SmolSent (Translation)}} 
& \multicolumn{2}{c}{\textbf{MAFAND (News)}} 
& \multicolumn{2}{c}{\textbf{MCS-350 (Stories)}} 
& \multicolumn{2}{c}{\textbf{\textsc{AfroScope} Test}} \\
& \multicolumn{2}{c}{$n = 54$} 
& \multicolumn{2}{c}{$n = 117$} 
& \multicolumn{2}{c}{$n = 25$} 
& \multicolumn{2}{c}{$n = 53$} 
& \multicolumn{2}{c}{$n = 21$} 
& \multicolumn{2}{c}{$n = 141$} 
& \multicolumn{2}{c}{$n = 640$} \\
\cmidrule(lr){2-3} 
\cmidrule(lr){4-5} 
\cmidrule(lr){6-7} 
\cmidrule(lr){8-9} 
\cmidrule(lr){10-11} 
\cmidrule(lr){12-13} 
\cmidrule(lr){14-15}
& F\textsubscript{1}\,$\uparrow$ & FPR\,$\downarrow$ 
& F\textsubscript{1}\,$\uparrow$ & FPR\,$\downarrow$ 
& F\textsubscript{1}\,$\uparrow$ & FPR\,$\downarrow$ 
& F\textsubscript{1}\,$\uparrow$ & FPR\,$\downarrow$ 
& F\textsubscript{1}\,$\uparrow$ & FPR\,$\downarrow$ 
& F\textsubscript{1}\,$\uparrow$ & FPR\,$\downarrow$
& F\textsubscript{1}\,$\uparrow$ & FPR\,$\downarrow$ \\
\midrule
AfroLID & $61.96$ & $0.0015$ & $56.61$ & $0.0014$ & $90.51$ & $0.0010$ & $58.06$ & $0.0020$ & $79.77$  & $0.0017$ & $52.53$ & $0.0011$ & $69.04$ & $0.0004$ \\
GlotLID-M & $95.25$ & $0.0003$ & $74.01$ & $0.0007$ & $94.33$ & $0.0003$ & $80.21$ & $0.0006$ & \textcolor{red}{\textbf{90.87}} & $0.0006$ & $76.95$ & $0.0002$ & $71.85$ & $0.0003$ \\
ConLID & $95.35$ & $0.0003$ & $74.76$ & $0.0008$ & $94.23$ & $0.0002$ & $79.58$ & $0.0006$ & $90.65$ & $0.0005$ & \textcolor{red}{\textbf{77.56}} & $0.0002$ & $72.90$ & $0.0002$ \\
OpenLID & $83.30$ & $0.0013$ & $26.19$ & $0.0034$ & $90.94$ & $0.0009$ & $47.31$ & $0.0031$ & $82.37$ & $0.0017$ & $39.83$ & $0.0020$ & $20.13$ & $0.2100$ \\
AfroLID\_\textit{AS} & $62.11$ & $0.0024$ & $59.74$ & $0.0020$ & $91.13$ & $0.0007$ & $64.75$ & $0.0024$ & $79.63$ & $0.0015$ & $54.70$ & $0.0011$ & $74.38$ & $0.0003$ \\
Serengeti\_\textit{AS} & $52.35$ & $0.0037$ & $57.72$ & $0.0023$ & $92.22$ & $0.0007$ & $60.05$ & $0.0034$ & $73.28$ & $0.0027$ & $45.70$ & $0.0016$ & $75.07$ & $0.0003$ \\
Cheetah\_\textit{AS} & $58.49$ & $0.0017$ & $62.34$ & $0.0016$ & $86.89$ & $0.0005$ & $59.67$ & $0.0018$ & $69.25$ & $0.0017$ & $51.53$ & $0.0008$ & $79.49$ & $0.0003$ \\
AfroLID\_\textit{GL} & $94.61$ & $0.0003$ & $72.35$ & $0.0009$ & $95.70$ & $0.0004$ & $84.08$ & $0.0007$ & $83.69$ & $0.0011$ & $73.39$ & $0.0006$ & $70.09$ & $0.0004$ \\
Serengeti\_\textit{GL} & $95.80$ & $0.0003$ & $72.82$ & $0.0009$ & $96.39$ & $0.0004$ & $84.48$ & $0.0008$ & $83.98$ & $0.0012$ & $75.55$ & $0.0005$ & $69.24$ & $0.0004$ \\
Cheetah\_\textit{GL} & $92.04$ & $0.0004$ & $72.58$ & $0.0008$ & $91.81$ & $0.0004$ & $78.81$ & $0.0008$ & $79.77$ & $0.0012$ & $68.65$ & $0.0006$ & $75.72$ & $0.0004$ \\
AfroLID\_\textit{AF} & $95.06$ & $0.0004$ & $74.61$ & $0.0009$ & $96.51$ & $0.0004$ & $85.07$ & $0.0007$ & $87.66$ & $0.0010$ & $74.56$ & $0.0005$ & $95.82$ & $0.0001$ \\
\quad + \textit{Mirror} & $96.29$ & $0.0003$ & $77.53$ & $0.0007$ & $95.47$ & $0.0004$ & $85.07$ & $0.0007$ & $88.69$ & $0.0009$ & $74.56$ & $0.0004$ & $97.74$ & $0.0000$ \\
Serengeti\_\textit{AF} & $95.04$ & $0.0004$ & $74.63$ & $0.0009$ & $96.51$ & $0.0004$ & $85.06$ & $0.0007$ & $87.66$ & $0.0010$ & $74.56$ & $0.0005$ & $95.82$ & $0.0001$ \\
\quad + \textit{Mirror} & \textbf{97.44} & $0.0003$ & \textbf{77.91} & $0.0006$ & \textbf{96.72} & $0.0004$ & \textbf{85.70} & $0.0007$ & $90.13$ & $0.0008$ & $76.03$ & $0.0004$ & \textbf{97.87} & $0.0000$ \\
Cheetah\_\textit{AF} & $95.62$ & $0.0003$ & $76.29$ & $0.0008$ & $93.70$ & $0.0003$ & $85.59$ & $0.0007$ & $85.42$ & $0.009$ & $71.77$ & $0.0005$ & $96.90$ & $0.0000$ \\
\bottomrule
\end{tabular}%
}

\caption{Language identification performance across seven evaluation benchmarks, where \textit{n} denotes the number of African languages in each benchmark. We report macro-F\textsubscript{1} and false positive rate (FPR), with higher macro-F\textsubscript{1} and lower FPR indicating better performance. Subscripts denote the fine-tuning corpus: \textsubscript{\textit{GL}} = GlotLID, \textsubscript{\textit{AS}} = AfroSimba, and \textsubscript{\textit{AF}} = \textsc{AfroScope}. \textit{+ Mirror} adds \textsc{AfroScope-Mirror} for targeted disambiguation. \linecolor{red}{Red underlines} indicate benchmark results affected by high training--evaluation contamination, and \textbf{bold} denotes  the best macro-F\textsubscript{1} score for each benchmark.}
\label{tab:better_eval}
\end{table*}

%% file: sections/experimental_setup.tex
\section{Experimental Setup}\label{sec:experimental_setup}

\paragraph{Evaluation data.} 
We evaluate on six external \textsc{Evaluation} sets spanning diverse domains and language coverage: \texttt{FLORES+}~\cite{nllb-24}, \texttt{UDHR}~\cite{kargaran-etal-2023-glotlid}, \texttt{MAFAND}~\cite{adelani-etal-2022-thousand}, \texttt{SmolSent}~\cite{caswell2025smol}, \texttt{MCS-350}~\cite{agarwal2023limit}, and \texttt{CommonLID}~\cite{suarez2026commonlid}. We also evaluate on a held-out \ourdata{} test split. Together, these benchmarks allow us to assess broad-coverage African LID and robustness across domains.

\paragraph{Contamination analysis.}
Because several African LID resources are derived from overlapping public sources, we explicitly measure residual overlap between \textsc{Train} and \textsc{Evaluation} data. We use a strict 4-gram contamination criterion: a test sentence is marked as contaminated if all of its 4-grams appear within a single training sentence. Table~\ref{tab:contamination} reports contamination rates across three training corpora and the six external evaluation sets. \ourdata~exhibits no detected contamination on any of the \textsc{Evaluation} sets under this criterion, whereas the AfroSimba and GlotLID-C corpora show non-trivial overlap, particularly on MCS-350 ($14.15$\% with GlotLID-C) and MAFAND ($8.13$\% with GlotLID-C). We return to the relationship between contamination and downstream model performance in~\S\ref{sec:result}.
\input{tables/data_contamination}

\subsection{Evaluation Protocol and Metrics}
Following~\cite{kargaran-etal-2023-glotlid}, we evaluate each model on the full benchmark rather than filtering examples to the model's supported language set. This setting reflects more realistic LID deployment conditions, where systems encounter text in languages outside their training label space and must avoid incorrectly assigning such text to an in-scope language. It also makes coverage differences visible: models with narrower label spaces are penalized when they force unsupported languages into related or superficially similar supported labels.

We report \textbf{macro-F\textsubscript{1}}, which averages performance across languages, together with \textbf{False Positive Rate} (FPR), defined as: $\text{FPR} = \frac{\text{FP}}{\text{FP} + \text{TN}}$, where FP is the number of false positives and TN is the number of true negatives. Macro-F\textsubscript{1} measures average per-language identification quality, while FPR captures how often a model falsely attributes text to a language it does not belong to. This distinction is important for African LID, where unsupported or closely related languages can otherwise be misclassified as higher-resource relatives.

\subsection{Baselines}
We compare against a diverse set of LID systems, ranging from FastText-based classifiers to transformer-based African language models.

\paragraph{FastText models.}
We evaluate three existing FastText-based LID systems: \textit{GlotLID-M}~\cite{kargaran-etal-2023-glotlid}, \textit{OpenLID}~\cite{burchell-etal-2023-open} and \textit{ConLID}~\cite{foroutan2025conlid}. \textit{ConLID} incorporates supervised contrastive learning to improve robustness on out-of-domain data.

\paragraph{Neural models.}
We evaluate three transformer-based models developed for African languages: \textit{AfroLID}~\cite{adebara2022afrolid}, \textit{Serengeti}~\cite{adebara2022serengeti}, and \textit{Cheetah}~\cite{adebara2024cheetah}. To isolate the effect of training data, we fine-tune these models on each \textsc{Train} source where applicable, and refer to the resulting models collectively as \ourmodel. Because \textit{SimbaText} is substantially smaller than the other training sources, we merge it with \textit{AfroLID} to form a combined training source, \textit{AfroSimba}. Fine-tuning hyperparameters are provided in Appendix~\ref{appdx_sec:baseline_models}.

%% file: tables/data_contamination.tex
\begin{table}[!ht]
\centering
\resizebox{0.45\textwidth}{!}{%
\begin{tabular}{lrrrrrr}
\toprule
\multirow{2}{*}{\textbf{Eval Set}} & \multicolumn{2}{c}{\textbf{AfroSimba}} & \multicolumn{2}{c}{\textbf{Glotlid-C}} & \multicolumn{2}{c}{\textbf{\textsc{AfroScope-DATA}}} \\
\cmidrule(lr){2-3}\cmidrule(lr){4-5}\cmidrule(lr){6-7}
 & \textbf{Cont.} & \textbf{Lang.} & \textbf{Cont.} & \textbf{Lang.} & \textbf{Cont.} & \textbf{Lang.} \\
\midrule
FLORES+    & $0.00\%$  & $0$   & $0.00\%$  & $0$  & $0.00\%$  & $0$ \\
UDHR       & $\mathbf{1.75\%}$  & $33$  & $0.97\%$  & $10$ & $0.00\%$  & $0$ \\
CommonLID  & $1.41\%$  & $16$  & $\mathbf{7.19\%}$  & $18$ & $0.00\%$  & $0$ \\
SMOL       & $0.00\%$  & $1$   & $\mathbf{0.04\%}$  & $16$ & $0.00\%$  & $0$ \\
MAFAND     & $0.37\%$  & $12$  & $\mathbf{8.13\%}$  & $18$ & $0.00\%$  & $0$ \\
MCS-350    & $1.47\%$  & $84$  & $\mathbf{14.15\%}$ & $97$ & $0.00\%$  & $0$ \\
\bottomrule
\end{tabular}%
}
\caption{Training--evaluation contamination across external benchmarks. We report the percentage of contaminated evaluation sentences (\textit{Cont.}) and the number of affected languages (\textit{Lang.}) per training corpus. Highest contamination rates are bolded.}
\label{tab:contamination}
\end{table}


%% file: sections/eval_results.tex
\section{Results}\label{sec:result}
Table~\ref{tab:better_eval} reports macro-F\textsubscript{1} and FPR across all external \textsc{Evaluation} sets for the full set of baselines and \ourmodel\ variants. \ourmodel\ achieve the strongest overall performance, leading on five of seven benchmarks, with the largest gains appearing on benchmarks with broader language coverage. The strongest improvements occur on the two broadest evaluation sets, \ourdata\ Test ($640$ languages) and UDHR ($117$ languages), where the best \textit{Serengeti\textsubscript{AF}\textsubscript{$+$ Mirror}} variant obtains $97.87$ and $77.91$ macro-F\textsubscript{$1$}, respectively, with FPRs of $0.0000$ and $0.0006$. Datasets with narrower language coverage, such as FLORES+, CommonLID, and SmolSent, show the same general pattern but with smaller margins. GlotLID-M and ConLID, both trained on the GlotLID-C corpus, achieve the strongest results only on MAFAND ($90.87$ and $90.65$ macro-F\textsubscript{1}) and MCS-350 ($77.56$ macro-F\textsubscript{1}). These are also the evaluation datasets for which the GlotLID-C  corpus exhibits the highest contamination, suggesting that evaluation overlap may contribute to their advantage on these benchmarks rather than reflecting  broader generalization.

\paragraph{Performance and coverage.}
Coverage is a major driver of performance differences across evaluation sets. Languages absent from a model's training label space are vulnerable to forced misattributions, where the model assigns them to a related or superficially similar supported language, thereby inflating FPR. Across the seven evaluation sets, the GlotLID-C corpus label space lacks coverage for $171$ evaluation-language entries while \ourdata\ covers all but $31$. On datasets where both corpora cover nearly all evaluation languages, such as FLORES+, MAFAND, and SmolSent, model performance clusters within a relatively narrow macro-F\textsubscript{1} range. In contrast, on broader evaluation sets with larger coverage gaps, such as UDHR and MCS-350, \ourmodel\ variants achieve an average macro-F\textsubscript{1} of $75.25$ compared to $67.25$ for non-\ourdata\ baselines, while average FPR decreases from $0.00101$ to $0.00067$.

\input{tables/open_source}

\begin{figure}[!ht]
\centering
\includegraphics[width=.99\columnwidth]{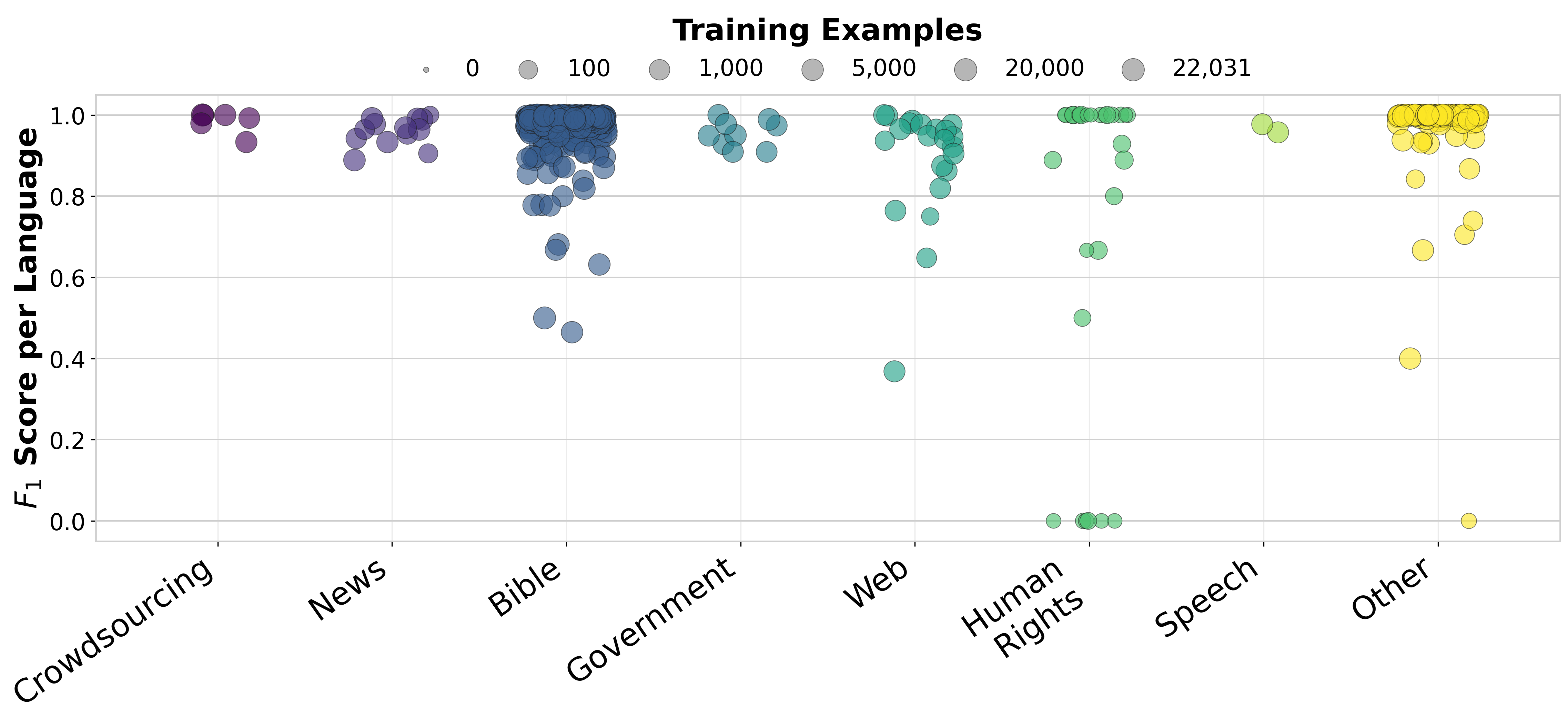}
\caption{Per-language macro-F\textsubscript{1} scores across domains. Bubble size corresponds to training examples.}
\label{domain_performance}
\end{figure}

\paragraph{Performance by domain.} \label{sec:domain_result}
Figure~\ref{domain_performance} shows that model performance is strongly shaped by the amount and distribution of domain-specific training data. High-resource domains such as \textit{Bible, News, Crowdsourcing,} and \textit{Government} generally achieve strong per-language macro-F\textsubscript{1} scores, with most languages clustered near the top of the distribution. In contrast, lower-resource or more heterogeneous domains exhibit weaker and less stable behavior. \textit{Human Rights} contains only a few training examples across $60$ languages, and its per-language results include a severe lower tail, with several languages receiving near-zero macro-F\textsubscript{1}. This pattern helps explain the relatively low performance on UDHR, a human-rights benchmark. Similarly, the \textit{Web} domain has moderate training coverage but noticeably weaker lower-tail performance, suggesting that domain diversity and text heterogeneity also affect robustness beyond raw example count alone. Overall, these results show that LID performance depends on both language coverage and domain coverage: domains that are well represented in training generalize more reliably, while sparse or unevenly distributed domains produce brittle performance for some languages.

\paragraph{Comparison with frontier LLMs.} Given that LLMs now set the state of the art across many NLP tasks, we evaluate three frontier LLMs on \ourdata\ test: Gemini-Pro, Claude-4.7-OPUS, and GPT-5.4-mini. Prompt templates, candidate-label formatting, and other evaluation details are provided in Appendix~\ref{appdx_sec:openllm_models}. As shown in Table~\ref{tab:open_source_results}, all three LLMs lag substantially behind dedicated LID systems: the strongest, Claude-4.7-OPUS, reaches only $47.40$ F\textsubscript{1} (FPR $0.0041$), compared to $69.08$ for GlotLID-M and $97.71$ for \ourmodel. This gap suggests that despite their broad multilingual capabilities, general-purpose LLMs are not a substitute for purpose-built LID models on African languages.

%% file: tables/open_source.tex
\renewcommand{\thefootnote}{\fnsymbol{footnote}}
\begin{table}[!ht]
\centering
\resizebox{0.4\textwidth}{!}{%
\begin{tabular}{ccc}
\toprule
\multirow{2}{*}{\textbf{Models}} & \multicolumn{2}{c}{\textbf{\textsc{AfroScope} Test\footnotemark}} \\
& $F1\uparrow$ & $FPR\downarrow$ \\
\midrule
GlotLID-M & $69.08$ & $0.0002$ \\
\textit{Serengeti}\textsubscript{\textit{AF}} & $97.71$ & $0.0001$ \\
\midrule
Gemini-Pro & $36.40$ & $0.0069$ \\
GPT-5.4-mini & $45.30$ & $0.0043$ \\
Claude-4.7-OPUS & $47.40$ & $0.0041$ \\
\bottomrule
\end{tabular}%
}
\caption{Comparison of dedicated LID systems and frontier LLMs on \textsc{AfroScope} Test.}
\label{tab:open_source_results}
\end{table}
\footnotetext{We downsample to 100 examples per language to reduce inference cost.}
\renewcommand{\thefootnote}{\arabic{footnote}}

%% file: sections/discussions.tex
\section{Analysis}\label{sec:discussion}
\noindent In this section, we analyze the patterns behind the aggregate results in Table~\ref{tab:better_eval}. We focus primarily on \ourdata~test because it has the broadest language coverage, while using external evaluation sets to test whether the same patterns recur across domains and benchmarks. Our analysis addresses two questions: \textit{(i)} how does confusability among closely related languages affect performance? and \textit{(ii)} which languages synergize or interfere most with one another, and how do language family relationships and script compatibility shape these effects?

\subsection{Confusability in related languages}


We find that \textit{label confusability} is a primary failure mode: the repeated false-positive assignment between a focus language and a small set of related or geographically proximate labels. Table~\ref{tab:low-f1-macro-confusions} previews representative confusion patterns identified across the evaluation sets. Such cases are especially common across ISO macrolanguage groupings and closely related languages or varieties that share substantial lexical and orthographic overlap. We find that these confusions recur in at least four of six evaluation sets, indicating the errors are stable across datasets rather than isolated benchmark artifacts. Among \textit{macrolanguage} varieties, Arabic (\textit{ara}: \textit{aeb, ary, arz}) and Fulah (\textit{ful}: \textit{fub, fuf, fuv}) are representative. High-confusion pairs are often genetically related and geographically proximate, such as Lamnso (\textit{lns}) and Tikar (\textit{tik}) in Cameroon, and Anii (\textit{blo}) and Lukpa (\textit{dop}) in Togo. These groupings account for a disproportionate share of false positives, suggesting that models struggle to discriminate fine-grained varieties. Together, these patterns indicate that many errors reflect genuine linguistic similarity and label granularity rather than random noise. We provide further examples in Appendix~\ref{app:more_confusion_examples}.

\input{tables/sil}

\begin{figure}[!ht]
\centering
\includegraphics[width=.45\textwidth]{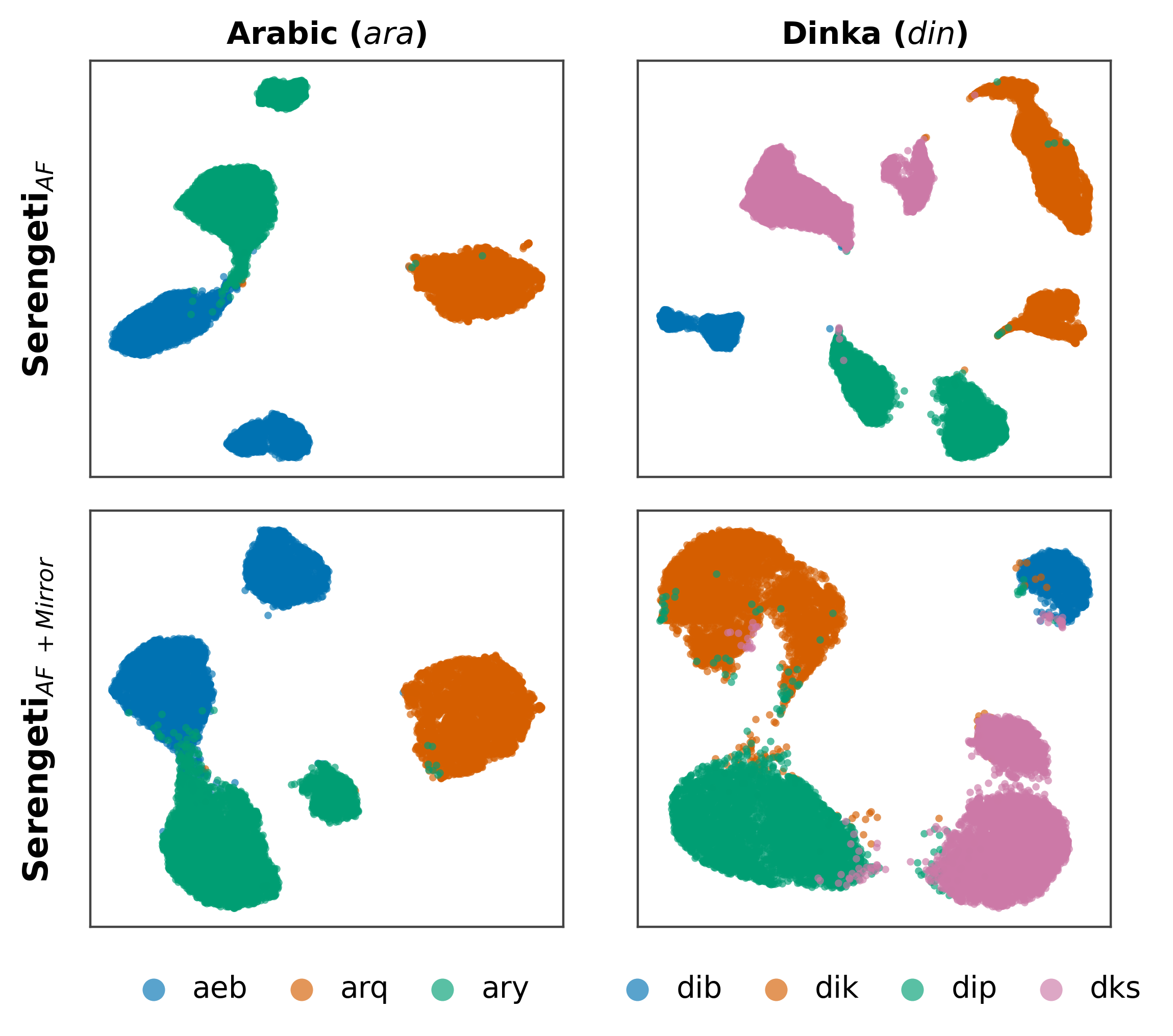}
\caption{UMAP of embedding spaces from base \textit{Serengeti\textsubscript{AF}} (top) and \textit{Serengeti\textsubscript{AF}\textsubscript{$+ Mirror$}} (bottom). We show examples from macro-language groups and high-confusion language pairs. Specialized embeddings exhibit improved separation between closely related varieties. We provide additional examples in Appendix~\ref{fig:umap_full}.}
\label{fig:umap_comparison}
\end{figure}


\subsubsection{\ourmirror{}   Disambiguation}\label{sec:hierarchical}

To address these systematic confusions, we introduce \ourmirror, a  lightweight contrastive embedding model used inside a hierarchical inference procedure. The base LID model first produces a broad-coverage prediction. When the prediction falls within a predefined confusable group and the model's confidence is below a threshold $\tau$, \ourmirror\ performs a group-specific disambiguation step using specialized embeddings. We build \ourmirror\ on top of AfroLID\textsubscript{\textit{AF}} and Serengeti\textsubscript{\textit{AF}}, our strongest base models on the external evaluation sets, and train it with Mirror-BERT~\cite{liu2021fast}, an unsupervised contrastive learning objective that pulls semantically similar representations together while pushing unrelated ones apart. Detailed training procedures and hyperparameters are in Appendix~\ref{tab:mirrorbert_hyperparams}.

Figure~\ref{fig:umap_comparison} compares the embedding spaces produced by Serengeti\textsubscript{\textit{AF}} and Serengeti\textsubscript{\textit{AF}} \textit{\textsubscript{$+$ Mirror}}, visually illustrating clearer separation among confusable labels and tighter within-label clustering after specialization. 

We evaluate this strategy on the $30$ confusable languages in Table~\ref{tab:full_hier}, including focus languages and their top confusion partners, across confidence thresholds. On the full set, the average macro-F\textsubscript{1} gain is +$1.57$ points. As shown in Table~\ref{tab:better_eval}, adding \ourmirror{} yields consistent gains across the external benchmarks: Serengeti\textsubscript{\textit{AF}} \textit{\textsubscript{$+$ Mirror}}{} improves over its base model on all seven evaluation sets (e.g., +$2.40$ on FLORES+, +$3.28$ on UDHR, +$2.47$ on Mafand), and AfroLID\textsubscript{\textit{AF}} \textit{\textsubscript{$+$ Mirror}}{} improves on five of seven datasets. At the language level (Table~\ref{tab:hier_results}), we observe improvements for closely related varieties within macro-language groupings, such as  Arabic (\textit{ara}) varieties \textit{aeb}, \textit{arq}, and \textit{ary}, which improve by +$0.63$, +$0.79$, and +$0.31$, respectively, as well as Swahili (\textit{swa}) variety \textit{swc} (+$0.55$). We also observe improvements for confusable regional pairs, including Kinyarwanda (\textit{kin}) and  \textit{run} (+$0.71$). However, a small number of languages decline, including \textit{niq} ($-0.10$) and \textit{kbp} ($-0.14$), suggesting that hierarchical routing can introduce unnecessary complexity when the base model is already reliable for a given label.

\input{tables/hier_results}

\noindent\subsection{Transfer Synergy and Interference}\label{sec:transfer}
Given limited and skewed resources, effective African LID depends on cross-lingual transfer. We investigate which languages synergize or interfere most, and how language family and script compatibility shape these effects. To quantify this, we adapt the Bilingual Transfer Score (BTS)~\cite{longpre2025atlas} to language identification. We compute BTS using macro-F\textsubscript{1}, and cross-entropy loss across $30$ representative African languages spanning five family clusters.

Figure~\ref{fig:family_matrix} shows that cross-lingual transfer in African LID is governed by shared script and fine-grained language relatedness rather than coarse family membership. Co-training yields the largest F\textsubscript{1} gains for low-resource targets with weak monolingual baselines (e.g., \textit{fuv}, \textit{dyu}, \textit{bam}; F\textsubscript{1} BTS up to +$0.22$), consistent with baseline rescue rather than family-specific transfer. Critically, the closest relationships are not uniformly beneficial: mutually intelligible varieties such as the Arabic cluster (\textit{arz}, \textit{ary}, \textit{aeb}) and Mande pair \textit{bam}--\textit{dyu} improve F\textsubscript{1} but \emph{degrade} probability calibration, with loss-based BTS as low as -$0.24$. A regression across all pairs confirms that script and this confusable-cousin relationship—not language family—are the dominant predictors of transfer; family membership alone is not significant.  This accuracy--calibration trade-off motivates hierarchical disambiguation. We report full per-family analysis in Appendix~\ref{app:transfer_synergy}.

%% file: tables/sil.tex
\begin{table}[!ht]
\centering
\resizebox{0.40\textwidth}{!}{%
\begin{tabular}{@{}cllcc@{}}
\toprule
& \textbf{Lang.} & \textbf{FP.} & \textbf{FC.} & \textbf{TC.} \\
\midrule
\multirow{3}{*}{\rotatebox{90}{\makecell[c]{\underline{\textbf{Macro.}}}}}
 & \textbf{ara} & $0.093$ & $2281$ & aeb, apd, arq, ary \\
 & \textbf{din} & $0.040$ & $497$ & dib, dik, dip, dks \\
 & \textbf{ful} & $0.010$ & $414$ & ffm, fub, fuc, fue \\
\midrule
\multirow{3}{*}{\rotatebox{90}{\makecell[c]{\underline{\textbf{High.}}}}}
 & \textbf{lns} & $0.281$ & $2577$ & tik, pfe, agq \\
 & \textbf{blo} & $0.199$ & $1748$ & kbp, dop, bib \\
 & \textbf{kin} & $0.153$ & $4323$ & run, swh, nnb \\
\bottomrule
\end{tabular}%
}
\caption{Representative false-positive patterns for highly confusable languages. We group cases into macrolanguage-related confusions and other high-confusion language groups. \textit{FP} reports the false positive rate (\%), \textit{FC} reports the false positive count, and \textit{TC} lists the most frequent target labels to which the focus language is incorrectly assigned.}
\label{tab:low-f1-macro-confusions}
\end{table}

%% file: tables/hier_results.tex
\begin{table}[!ht]
\centering
\resizebox{\columnwidth}{!}{%
\scriptsize
\renewcommand{\arraystretch}{1.5}
\begin{tabular}{l c r r r r r r r}
\toprule
\multirow{2}{*}{\textbf{Group}} & \multirow{2}{*}{\textbf{Lang}} & \multirow{2}{*}{\textbf{Baseline}}
& \multicolumn{2}{c}{$\tau=0.75$} & \multicolumn{2}{c}{$\tau=0.85$} & \multicolumn{2}{c}{$\tau=0.95$} \\
\cmidrule(lr){4-5} \cmidrule(lr){6-7} \cmidrule(lr){8-9}
& & & $\boldsymbol{\Delta}$ & \textbf{F\textsubscript{1}} & $\boldsymbol{\Delta}$ & \textbf{F\textsubscript{1}} & $\boldsymbol{\Delta}$ & \textbf{F\textsubscript{1}} \\
\midrule
\multirow{3}{*}{ara} 
& aeb & $94.50$ & $+0.18$ & $94.68$ & $\mathbf{+0.63}$ & $\mathbf{95.13}$ & $+0.50$ & $95.00$ \\
& arq & $98.89$ & $+0.32$ & $99.21$ & $+0.64$ & $99.53$ & $\mathbf{+0.79}$ & $\mathbf{99.68}$ \\
& ary & $96.05$ & $+0.11$ & $96.16$ & $\mathbf{+0.31}$ & $\mathbf{96.35}$ & $+0.13$ & $96.17$ \\
\midrule
\multirow{1}{*}{swa} 
& swc & $94.21$ & $+0.34$ & $94.55$ & $+0.07$ & $94.28$ & $\mathbf{+0.55}$ & $\mathbf{94.77}$ \\
\midrule
\multirow{1}{*}{kin} 
& run & $95.34$ & $+0.03$ & $95.37$ & $+0.03$ & $95.37$ & $\mathbf{+0.71}$ & $\mathbf{96.05}$ \\
\midrule
\textbf{Avg} & & -- & $+0.19$ & -- & $+0.33$ & -- & $\mathbf{+0.54}$ & -- \\
\bottomrule
\end{tabular}
}
\caption{Sample hierarchical classification results using \ourmirror~embeddings across confidence thresholds $\tau$. Baseline F\textsubscript{1} shows base Serengeti performance; $\boldsymbol{\Delta}$ columns show improvement in F\textsubscript{1} over baseline. Bold indicates the best F\textsubscript{1} per language. Full results are provided in Appendix~\ref{tab:full_hier} }
\label{tab:hier_results}
\end{table}

%% file: sections/conclusion.tex
\section{Conclusion}\label{sec:conc}
We introduce \ourframework, a unified framework for African LID that combines broad-coverage data, strong LID models, targeted disambiguation, and linguistic analysis. We present \ourdata, a large-scale dataset spanning \numlanguages~language labels, which we use to train \ourmodel, a family of African LID models that outperform prior African-focused  baselines across internal and external evaluations. 

To address persistent confusions among closely related languages and varieties, we propose a hierarchical inference approach based on \ourmirror, a specialized embedding model for frequently confused language groups. On the full set of identified confusion groups, this approach improves macro-F\textsubscript{1} by $+1.57$ points on average.

Finally, our transfer analysis shows that African LID transfer is shaped by script, fine-grained family relatedness and confusability than by coarse family membership. These relationships can improve low-resource performance while degrading calibration, motivating targeted disambiguation with \textsc{AfroScope-Mirror}. We hope \ourframework, \ourdata, and \ourmodel\ support reliable African NLP and future work on fine-grained varieties, domain shifts, and mixed-language text.

%% file: sections/limit_ethics.tex
\section*{Limitations}\label{sec:limits}

We note several limitations of the current study.

\begin{enumerate}
    \item \textbf{Mixed-language and code-switched text.}
    Our formulation treats each instance as belonging to a single language label. This does not fully capture important phenomena in Africa's linguistic landscape, including code-switching, mixed-language documents, translanguaging practices, and contact varieties such as pidgins and creoles. Extending \ourframework\ to multi-label, document-level, or span-level language identification is an important direction for future work. Relatedly, the granularity at which languages are annotated is uneven across our data sources, and in some cases closely related varieties or distinct languages may be conflated under a single label. A more careful audit of these labeling decisions, ensuring that language distinctions are drawn consistently and appropriately, is needed but lies beyond the scope of this work.

    \item \textbf{Language metadata and classification choices.}
    \ourdata\ relies on external catalogs, primarily Ethnologue, to assign language identifiers, genealogical groupings, and script metadata. Alternative resources, such as Glottolog, may differ in classification, naming, macrolanguage treatment, and family hierarchy. These choices may affect analyses that depend on genealogical proximity or script compatibility. Future work should evaluate sensitivity to alternative metadata sources and provide mappings across catalog standards.

    \item \textbf{Confidence-based routing and calibration.}
    Our hierarchical disambiguation method relies on model confidence to decide when to invoke the group-specific refinement step. While this improves performance on many confusable labels, gains are not uniform across languages or thresholds, and some cases exhibit degradation. Improving probability calibration and learning a routing policy, rather than relying on fixed confidence thresholds, may further increase robustness.
    
   \item \textbf{Domain and source imbalance.}
    Although \ourdata\ improves domain diversity relative to prior African LID resources, some domains remain substantially smaller or more heterogeneous than others. As a result, performance may still be brittle for domains with limited training data or high internal variation, such as web text, human-rights documents, or informal user-generated content. Future work should expand naturally occurring data in underrepresented domains and evaluate robustness under domain shift.
\end{enumerate}

%% file: sections/appendix.tex
\clearpage
\appendix
\appendixpage            
\addappheadtotoc         
\numberwithin{figure}{section}
\numberwithin{table}{section}

The following appendices provide comprehensive supplementary material supporting the main findings of this work. We include an expanded literature review, detailed descriptions of the datasets and models, a more in-depth discussion of the experimental setup, and additional analyses and discussions.

\begin{itemize}
    \item \S\ref{appdx_sec:lit_review}: More in depth literature review
    \item \S\ref{app_sec:data_curation}: Data Curation \& Preprocessing
    \item \S\ref{appdx_sec:baseline_models}: Baseline Models
    \item \S\ref{appx:discussions}: Discussions
\end{itemize}

\input{appendix_sections/appdx_lit_review}
\input{appendix_sections/app_data_collection}

\input{appendix_sections/appdx_baseline_models}
\input{appendix_sections/appdx_discussion}


%% file: appendix_sections/appdx_lit_review.tex
\section{Literature Review}\label{appdx_sec:lit_review}
\subsection{Linguistic Diversity of Africa.}\label{appdx_sec:lit_review_diversity}
Many African languages are agglutinative (e.g., Swahili, Zulu), complicating tokenization and parsing~\citep{abdulmumin-etal-2024-correcting, hussen2025state}, while everyday communication often involves code-switching with English and other languages~\citep{ogunremi2023multilingual}.

Datasets like AfroCS-xs demonstrate that small, high-quality resources can boost LLM performance on code-switched tasks~\citep{olaleye-etal-2025-afrocs}. Models such as Afrolid, SERENGETI and Cheetah show the feasibility of massively multilingual modeling for African languages~\citep{adebara2022afrolid,adebara2022serengeti, adebara2024cheetah}. Benchmarks including Sahara~\cite{adebara2025we}, SimbaBench~\cite{elmadany2025voice}, IrokoBench~\cite{adelani2024irokobench}, and AfroBench~\cite{ojo2023good}. 

\subsection{Data Authenticity.}\label{appdx_sec:lit_review_authenticity}
Studies of multilingual corpora such as ParaCrawl~\citep{banon2020paracrawl}, WikiMatrix~\citep{schwenk2021wikimatrix}, and mC4~\citep{xue2021mt5} reveal widespread errors, mislabeled data, and ambiguous language codes~\citep{kreutzer2022quality}.
For instance, \citet{alabi-etal-2020-massive} showed that many FastText embeddings were heavily mislabeled with words from other languages. Poor data authenticity directly undermines cultural representation, as inaccurate or mislabeled data fails to capture the nuanced linguistic and cultural contexts inherent in low-resource languages~\citep{zhong2024opportunities}. Addressing these quality issues is essential for building reliable language technologies ~\citep{alhanai2025bridging, ojo2023good}.


%% file: appendix_sections/app_data_collection.tex
\section{Data Curation and Sampling}\label{app_sec:data_curation}
\subsection{Data Curation}\label{app_sec:data_curation_curation}
We deduplicate sentences across all sources to ensure that each example appears at most once in \ourdata. We then partition the data into \textit{train}, \textit{dev}, and \textit{test} splits, performing deduplication prior to splitting so that no sentence is shared across splits. We ensure that each sentence is written in the correct script, based on the writing system databases of Ethnologue~\cite{ethnologue2021}. When temperature sampling we set the maximum sentences to $10$M sentences. The resulting splits contain $5{,}952{,}573$ sentences for train, $463{,}875$ for dev, and $232{,}563$ for test. The full list of languages supported by \ourframework, together with the number of sentences per language, is provided in Tables~\ref{tab:train-balance-10m-languages-i}--\ref{tab:train-balance-10m-languages-iv}.

Below, we describe the sources used to construct \ourdata.

\paragraph{Train Corpus}
\paragraph{GlotLID-C.} We extract 523 African languages from \texttt{GlotLID-C}, a collection spanning 2,099 languages globally~\cite{kargaran-etal-2023-glotlid}.
\paragraph{AfroLID.} A manually curated multi-domain web dataset covering 513 African languages~\cite{adebara2022afrolid}.
\paragraph{SimbaText.} Speech-derived text data spanning 103 African languages, originally collected for speech and language identification~\cite{elmadany2025voice}.

\paragraph{Evaluation Datasets}
\paragraph{FLORES$+$.} Professionally translated sentences across $200$ languages, providing a clean and standardized benchmark, particularly for low-resource languages~\cite{nllb-24}.
\paragraph{MAFAND.} Manually audited data from news, Wikipedia, and religious texts across 55 African languages~\cite{adelani-etal-2022-thousand}.
\paragraph{SMOL.} Multilingual benchmark dataset consisting of professionally translated sentence- and document-level data for 115 LRLs~\cite{caswell2025smol}.
\paragraph{MCS-350.} Parallel children's stories across 151 African languages, drawn from a multilingual collection of 50k texts in over 350 languages~\cite{agarwal2023limit}.
\paragraph{UDHR.} The Universal Declaration of Human Rights, translated into a wide range of languages~\cite{kargaran-etal-2023-glotlid}.
\paragraph{CommonLID.} Professionally translated parallel data for 115 under-represented languages~\cite{suarez2026commonlid}.

\subsection{Domain classification.} \label{app_sec:domain_meta}
We assign domains by matching keywords found in the metadata associated with each sentence, following the categorization scheme from~\cite{kargaran-etal-2023-glotlid}. 
Table~\ref{tab:domain_mapping} lists the specific keyword mappings.

\input{tables/metadata}
\input{tables/domain_mapping}

%% file: tables/metadata.tex
\begin{table}[!h]
    \centering
    \small
    \begin{tabular}{l}
        \toprule
        \textbf{Sample Metadata} \\
        \midrule
        \texttt{Bible-aar\_line94} \\
        \texttt{CommonVoice v11} \\
        \texttt{Masakhanews} \\
        \texttt{GlotStoryBook} \\
        \texttt{CC100\_zu.txt.tsv\_f17\_line40162} \\
        \texttt{JW-zul\_line3295} \\
        \texttt{gov-za} \\
        \texttt{Open Subtitles} \\
        \texttt{KDE4} \\
        \texttt{Vuk'uzenzele} \\
        \bottomrule
    \end{tabular}
    \caption{Examples of sentence-level metadata identifiers used for domain attribution.}
    \label{tab:metadata_samples}
\end{table}

%% file: tables/domain_mapping.tex
\begin{table}[!t]
\centering
\scriptsize
\setlength{\tabcolsep}{3pt}
\renewcommand{\arraystretch}{1.1}

\resizebox{0.85\columnwidth}{!}{%
    \begin{tabularx}{\linewidth}{l X}
    \toprule
    \textbf{Domain} & \textbf{Associated Keywords} \\
    \midrule
    \textbf{Speech} & Speech, CommonVoice, TTS, Audio \\
    \textbf{Human Rights} & Human Rights \\
    \textbf{Government} & Human Rights, Autshumato, Legal, GOV, Parliament, Gazette \\
    \textbf{CrowdSource} & Flores, NLB, mt560, Tatoeba, UD, ai4d, lti, Benchmark, Human, Madar, iadd \\
    \textbf{News} & News, xlsum, Vukuzenzele, CBC, BBC, Afriqa, Masakha, Goud \\
    \textbf{Bible} & Bible, JW,  Scripture, Religion \\
    \textbf{Web} & Oscar, CC, CommonCrawl, Web, Dialect, Social, Forum \\
    \textbf{Other} & Health, Covid, Medical, Med, Tanzil, PBC, Quran, Story, Stories, Fiction, Bloom, Lyrics \\
    \bottomrule
    \end{tabularx}%
} 

\caption{Keywords extracted from dataset metadata to map sources into domain categories.}
\label{tab:domain_mapping}
\end{table}

%% file: appendix_sections/appdx_baseline_models.tex
\section{Training Hyperparameters for Baseline Models}\label{appdx_sec:baseline_models}
To isolate the effect of training data, we train or fine-tune comparable model architectures on each \textsc{Train} source where applicable. This allows us to evaluate whether the gains come from the model architecture, the training corpus, or their interaction. Because \textit{SimbaText} is substantially smaller than the other training sources, we merge it with \textit{AfroLID} to form a combined training source, \textit{AfroSimba}. AfroLID and Serengeti are XLM-RoBERTa variants, and Cheetah is a T5-based model.

\subsection{Neural Models} \label{app:baseline_neural}
We use the hyperparameters in Table~\ref{tab:neural_hyperparams} for \textit{AfroLID}~\cite{adebara2022afrolid} and \textit{Serengeti}~\cite{adebara2022serengeti} (both XLM-R variants), and Table~\ref{tab:cheetah_hyperparams} for \textit{Cheetah}~\cite{adebara2024cheetah}.

\section{Evaluating Open-Srouce LLMs}\label{appdx_sec:openllm_models}
\subsection{Prompts for Open-Source Models} \label{app:open_source}
Figure~\ref{fig:lid_prompt} shows the prompt template used to elicit language predictions from the LLMs. We provide the candidate ISO 639-3 labels in the prompt and instruct the model to return only the language code. To reduce inference cost we downsample each language examples to 100 examples per language.

\input{tables/neural_hyper}
\input{tables/cheetah_hyper}
\input{tables/prompt}

%% file: tables/neural_hyper.tex
\begin{table}[h]
    \centering
    \resizebox{0.85\columnwidth}{!}{%
        \begin{tabular}{llc}
            \toprule
            \textbf{argument} & \textbf{description} & \textbf{value} \\
            \midrule
            -max\_seq\_length & max input sequence length & 128 \\
            -per\_device\_train\_batch\_size & training batch size (per device) & 64 \\
            -learning\_rate & learning rate & 2e-5 \\
            -num\_train\_epochs & number of training epochs & 10 \\
            -metric\_for\_best\_model & evaluation metric & f1 \\
            \bottomrule
        \end{tabular}%
    }
    \caption{Training hyperparameters for Afrolid and Serengeti.}
    \label{tab:neural_hyperparams}
\end{table}

%% file: tables/cheetah_hyper.tex
\begin{table}[!h]
    \centering
    \resizebox{0.85\columnwidth}{!}{%
        \begin{tabular}{llc}
            \toprule
            \textbf{argument} & \textbf{description} & \textbf{value} \\
            \midrule
            -max\_target\_length & max target sequence length & 128 \\
            -per\_device\_train\_batch\_size & training batch size (per device) & 32 \\
            -learning\_rate & learning rate & 5e-5 \\
            -num\_train\_epochs & number of training epochs & 10 \\
            -metric\_for\_best\_model & evaluation metric & f1 \\
            \bottomrule
        \end{tabular}%
    }
    \caption{Training hyperparameters for the Cheetah model.}
    \label{tab:cheetah_hyperparams}
\end{table}

%% file: tables/prompt.tex
\begin{figure}[t]
\centering
\small
\setlength{\tabcolsep}{6pt}
\renewcommand{\arraystretch}{1.2}

\begin{tabular}{p{0.15\linewidth} p{0.78\linewidth}}
\toprule
\textbf{Field} & \textbf{Content} \\
\midrule

\textbf{Prompt} &
\ttfamily
Classify the language of the given sentence. Return only the ISO 639-3 language code from the list below. \\
& \ttfamily
Language options: aba, abi, abn, acd, ach, ada, \ldots, zul \\
& \ttfamily
Return only the language code, nothing else. \\

\midrule

\textbf{Input} &
\ttfamily
Sentence: \{sentence\} \\

\midrule

\textbf{Output} &
\ttfamily
amh \\

\bottomrule
\end{tabular}

\caption{Prompt template used for LLM-based language identification.}
\label{fig:lid_prompt}
\end{figure}

%% file: appendix_sections/appdx_discussion.tex
\section{Discussions}\label{appx:discussions}
\subsection{Confusability in related languages}\label{app:more_confusion_examples}
To investigate languages with high FPR, we isolate languages with scores below F\textsubscript{1} 90 across all evaluation datasets—and identify the top three most frequent misclassifications for each to form confusion groups. This analysis yields $13$ distinct groups comprising $30$ languages in total. We find that these confusions primarily stem from either macrolanguage structures (e.g., \textit{ful} vs. \textit{fub}) or geographic proximity (e.g., \textit{bsq} vs. \textit{bas}). Table~\ref{tab:full_hier} details the composition of all confusion groups, and we report the corresponding performance improvements for each individual language.

\begin{figure*}[!ht]
\centering
\includegraphics[width=.85\textwidth]{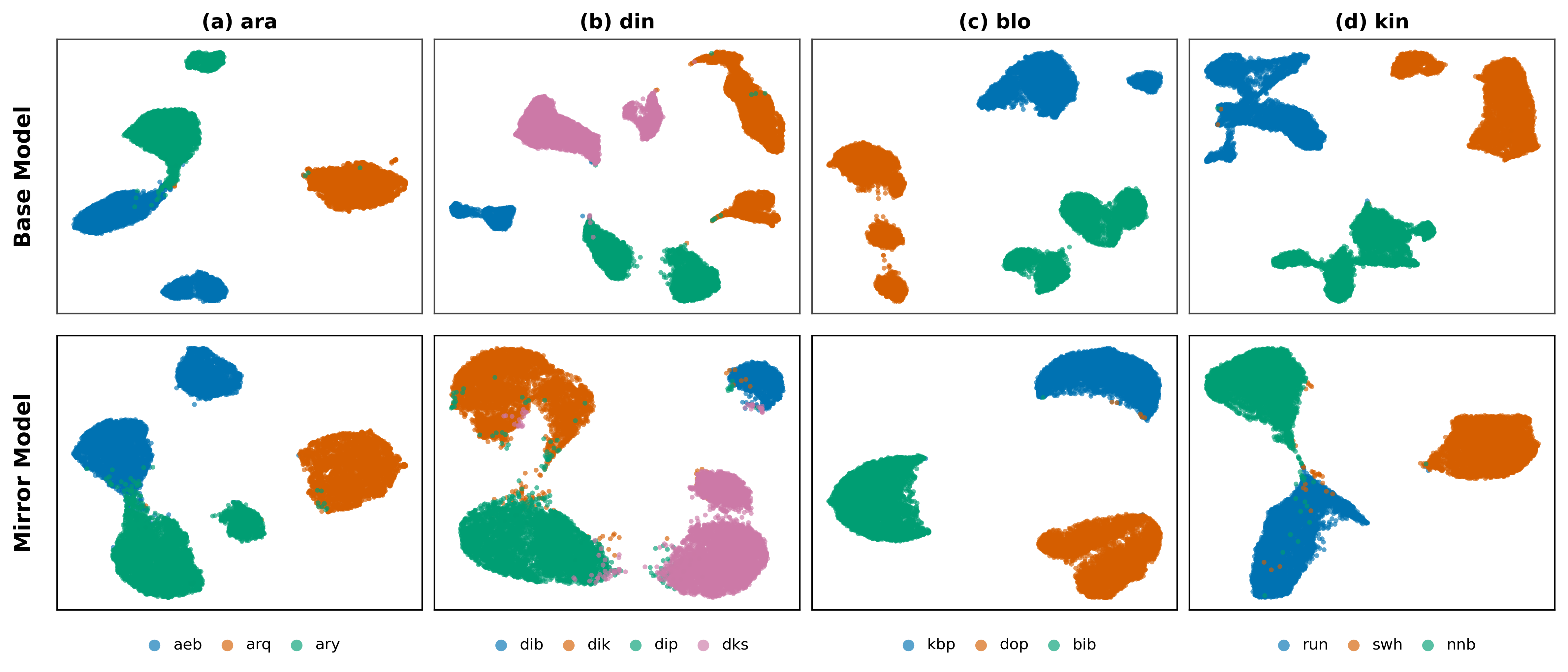}
\caption{UMAP visualization comparing base \textit{Serengeti\textsubscript{AF}} (top) and \textit{Serengeti\textsubscript{AF + Mirror}}(bottom) embedding spaces. We visualize four groups representing macro-languages and confusion pairs. Specialized embeddings show improved separation between closely related language varieties.}
\label{fig:umap_full}
\end{figure*}

\paragraph{Mirror-BERT training procedure}
We follow the procedures from the official github repository for Mirror-BERT\footnote{https://github.com/cambridgeltl/mirror-bert}. We use \textit{Serengeti\textsubscript{AF}} for this experiment as it being our best performing model. Table~\ref{tab:mirrorbert_hyperparams} shows the hyperparamerters we use to train \ourmirror. 
We see in Figure~\ref{fig:umap_full} that the label spaces of \textit{Serengeti\textsubscript{AF + Mirror}}~ shows much better separation in the embedding spaces than \textit{Serengeti\textsubscript{AF}}.

\input{tables/mirror_hyper}

\subsection{Transfer Synergy and Interference}\label{app:transfer_synergy}
We adapt the Bilingual Transfer Score (BTS)~\cite{longpre2025atlas} to language identification. For a source $s$ and target $t$, we compare a monolingual model trained on target data $\mathcal{D}_t$ against a bilingual model trained on $\mathcal{D}_t \cup \mathcal{D}s$, evaluated on a held-out target test set. For cross-entropy loss, where lower is better, we define: 
\begin{equation}
\mathrm{BTS}^{\mathcal{L}}_{s \rightarrow t}
=
\frac{
\mathcal{L}_t(\mathcal{D}_t) - \mathcal{L}_t(\mathcal{D}_t \cup \mathcal{D}_s)
}{
\mathcal{L}_t(\mathcal{D}_t) + \epsilon
}.
\label{eq:bts_loss}
\end{equation}
so that positive values indicate improvement. For macro-F\textsubscript{1}, where higher is better, we define: 
\begin{equation}
\mathrm{BTS}^{F1}_{s \rightarrow t}
=
\frac{
F1_t(\mathcal{D}_t \cup \mathcal{D}_s) - F1_t(\mathcal{D}_t)
}{
F1_t(\mathcal{D}_t) + \epsilon
}.
\label{eq:bts_f1}
\end{equation}

\subsubsection{Language selection}\label{app:transfer_selection}
\begin{figure*}[!ht]
\centering
\includegraphics[width=.85\textwidth]{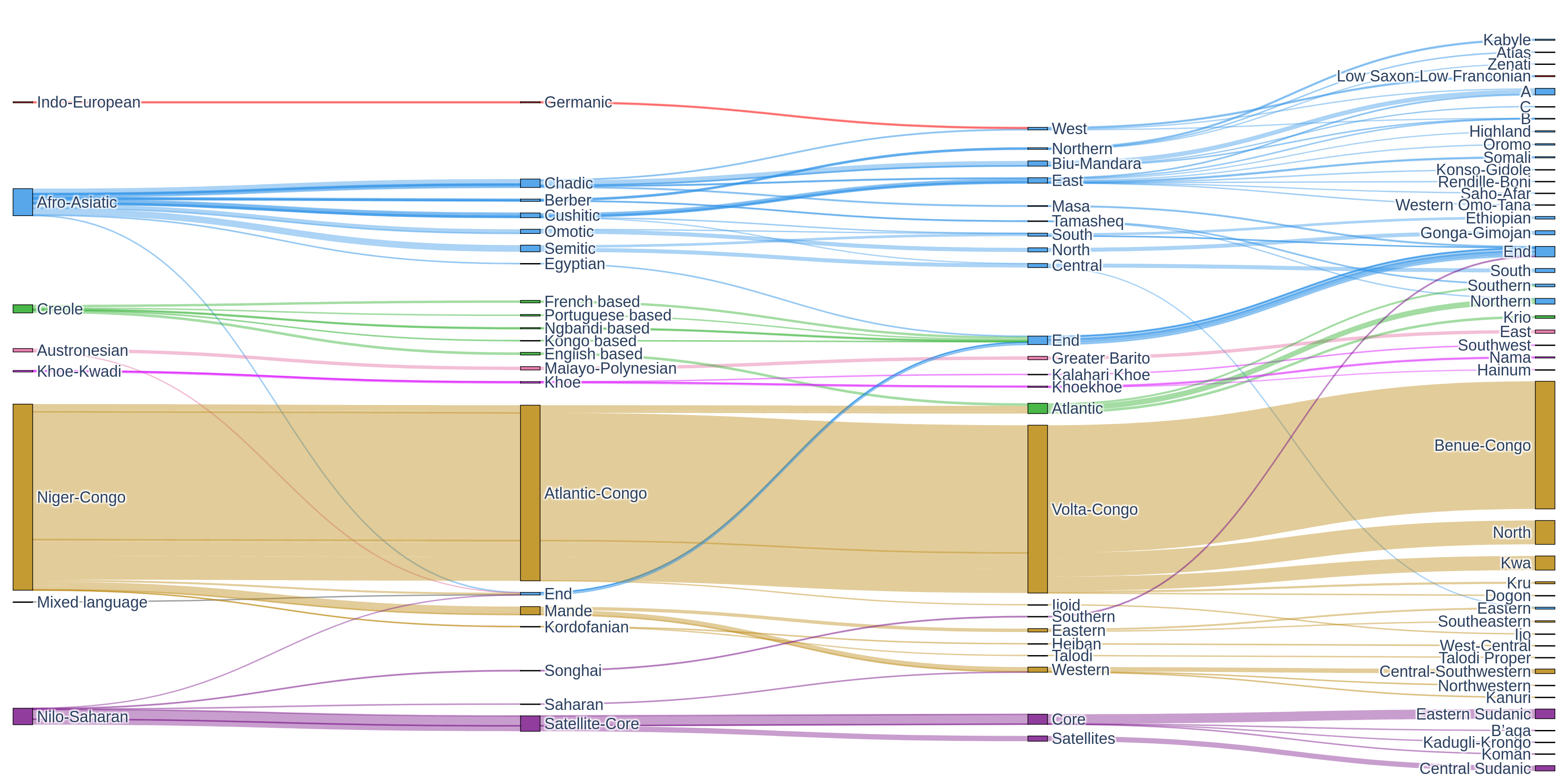}
\caption{Distribution of languages across major language groupings, intermediate sub-families, and finer-grained groupings, capturing their genetic relationships.}
\label{fig:lang_hier}
\end{figure*}

We select $30$ languages spanning five clusters. To populate a range of genealogical distances, we group eligible languages (those with at least \textsc{$10{,}000$} sentences) by their shared family hierarchy and retain clusters that contain documented confusable pairs, drawn from our false-positive analysis (Appendix~\ref{app:more_confusion_examples}). This yields clusters of mutually intelligible varieties—the Arabic dialects (\textit{arz}, \textit{ary}, \textit{aeb}), Fula varieties (\textit{fub}, \textit{fuv}, \textit{fuh}, \textit{fuq}, \textit{fue}), and the Mande pair (\textit{bam}, \textit{dyu})—alongside more distant same-family languages and unrelated outgroups (\textit{plt}, \textit{afr}, \textit{sag}). Figure~\ref{fig:lang_hier} shows the fine-grained relatedness of the selected languages across the family hierarchy.

\subsubsection{Experimental setup}\label{app:transfer_setup}
Each language contributes $1{,}000$ target sentences and a held-out test set; bilingual models receive $200$ background sentences from every other language so all models share one label space. We fine-tune \textit{Serengeti\textsubscript{AF}} for the $30$ monolingual baselines (median over three seeds) and the $435$ language pairs (one seed), yielding a $30$ X $30$ directional transfer matrix per metric.

\paragraph{F\textsubscript{1} gains track headroom.} The largest F\textsubscript{1}-BTS values occur for the weakest baselines—\textit{fuv} ($0.62$), \textit{fuq} ($0.53$), \textit{dyu} ($0.69$), \textit{bam} ($0.73$)—which improve from nearly any source (\textit{bam}$\rightarrow$\textit{dyu}~$=+0.22$, \textit{fub}$\rightarrow$\textit{fuv}~$=+0.16$ ; Table~\ref{tab:top_bts}). F\textsubscript{1}-BTS correlates with target headroom ($r=0.60$), indicating low-resource rescue rather than family-specific transfer; strong baselines (\textit{kin}, \textit{xho}, \textit{tso}, \textit{tsn}, all $>0.98$) show near-zero BTS.

\paragraph{Cousins degrade calibration.} The clearest signal is in loss-BTS. The Arabic varieties form a block of negative loss-BTS (\textit{aeb}$\rightarrow$\textit{ary}~$=-0.24$, \textit{aeb}$\rightarrow$\textit{arz}~$=-0.18$, \textit{ary}$\rightarrow$\textit{aeb}~$=-0.15$), and the Mande pair replicates this (\textit{dyu}$\rightarrow$\textit{bam}~$=-0.17$). All show \emph{positive} F\textsubscript{1}-BTS: co-training mutually intelligible varieties picks the right label but spreads probability mass across the cousin, raising cross-entropy. This trade-off is visible in Figure~\ref{fig:family_script_synergy}, where cousin pairs sit in the mutual-gain region under F\textsubscript{1} but the mutual-loss region under cross-entropy.  Bambara (\textit{bam}) and Jula (\textit{dyu}) are adjacent Manding varieties spoken across Mali, Côte d'Ivoire, and Burkina Faso~\cite{ethnologue2021}, so this interference is expected.

\paragraph{Script gates transfer.} Substantial transfer is confined to same-script pairs: $78\%$ of F\textsubscript{1} changes with $|\mathrm{BTS}|\geq0.05$ and all calibration changes with $|\mathrm{BTS}|\geq0.10$ occur within a script (Table~\ref{tab:script_pairs}). Controlling for headroom and cousin status, shared script independently predicts larger F\textsubscript{1}-transfer magnitude ($\beta=+0.005$). The exception is low-resource rescue, where weak Fula targets improve from any source. Distinctive scripts with strong baselines, like the Ethiopic pair \textit{amh}--\textit{tir} ($>0.99$), show negligible transfer (mean $|\mathrm{BTS}^{\mathcal{L}}|=0.006$), as orthographic distinctiveness leaves no confusion to resolve.

\paragraph{Distance gates the effect.} Different-family pairs show loss-BTS near zero (distance-$5$ mean~$=+0.002$, $n=472$). Regressing loss-BTS on genealogical distance $d$ and $d^2$ yields a significant inverted-U ($\beta_d=+0.015$, $\beta_{d^2}=-0.002$): distance-$0$ cousins are uniquely negative (mean~$=-0.029$), all greater distances near zero. F\textsubscript{1}-BTS shows no distance pattern, masked by ceiling saturation.

\input{tables/full_hierarchy}

\input{tables/tab_top_bts}
\input{tables/tab_script_pairs}

\begin{figure*}[!ht]
\centering
\includegraphics[width=.95\textwidth]{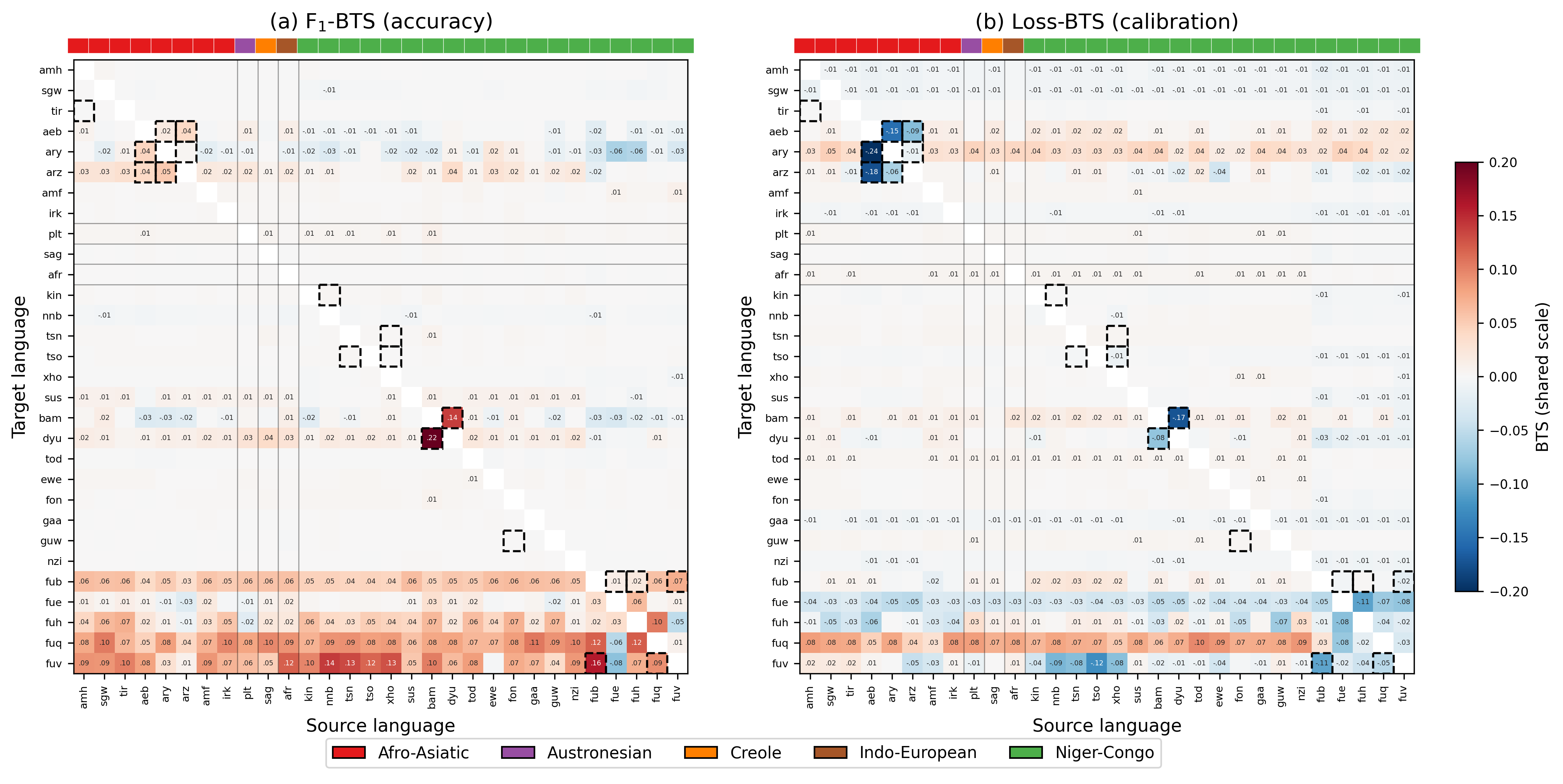}
\caption{Pairwise transfer effects across African languages, measured by \textbf{(a)} F$_1$-BTS (accuracy) and \textbf{(b)} loss-BTS (calibration). Rows indicate target languages and columns indicate source languages used for co-training; cell values show the transfer score, with warmer colors indicating positive transfer and cooler colors indicating negative transfer. Color bars along the axes indicate high-level language-family groupings. Dashed boxes highlight selected within-family or same-script clusters where transfer is especially structured, such as Arabic varieties, Southern Bantu languages, Mande languages, and Fulah varieties. Both panels share a common symmetric color scale; note that F$_1$-BTS and loss-BTS are distinct quantities, so the same pair may gain accuracy in (a) while losing calibration in (b).}
\label{fig:family_script_synergy}\label{fig:family_matrix}
\end{figure*}

\begin{figure*}[!ht]
\centering
\includegraphics[width=.95\textwidth]{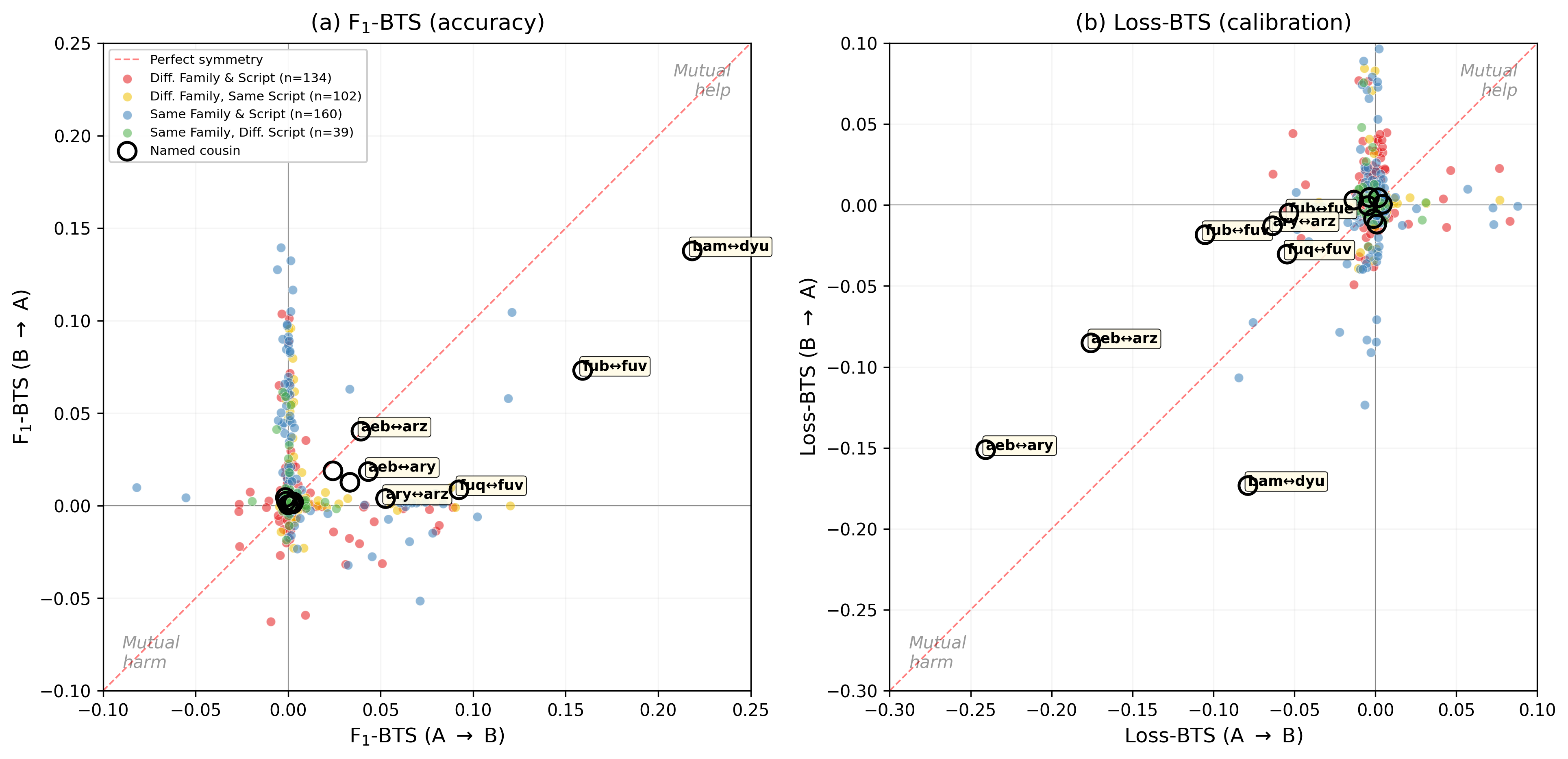}
\caption{Directional transfer symmetry for \textbf{(a)} F$_1$-BTS and
\textbf{(b)} loss-BTS. Each point is a language pair; axes show BTS in each
direction, and the dashed line marks perfect symmetry. Confusable cousins
(circled) sit in the mutual-gain (upper-right) region under F$_1$ but the
mutual-loss (lower-left) region under cross-entropy, illustrating the
accuracy--calibration trade-off; off-diagonal points indicate asymmetric
transfer.}
\label{fig:family_script_synergy}
\end{figure*}

\input{tables/supported_lang}

%% file: tables/mirror_hyper.tex
\begin{table}[h]
    \centering
    \resizebox{0.85\columnwidth}{!}{%
        \begin{tabular}{llc}
            \toprule
            \textbf{argument} & \textbf{description} & \textbf{value} \\
            \midrule
            -epoch & number of training epochs & 1 \\
            -train\_batch\_size & training batch size & 200 \\
            -learning\_rate & learning rate & 2e-5 \\
            -max\_length & max sequence length & 50 \\
            -infoNCE\_tau & InfoNCE temperature ($\tau$) & 0.04 \\
            -dropout\_rate & dropout rate & 0.0 \\
            -drophead\_rate & drophead rate & 0.05 \\
            -random\_span\_mask & length of random span mask & 5 \\
            -agg\_mode & aggregation mode & cls \\
            \bottomrule
        \end{tabular}%
    }
    \caption{Training hyperparameters for the Mirror-BERT model.}
    \label{tab:mirrorbert_hyperparams}
\end{table}

%% file: tables/full_hierarchy.tex
\begin{table*}[!ht]
\centering
\resizebox{.85\textwidth}{!}{%
\scriptsize
\renewcommand{\arraystretch}{1.4}
\begin{tabular}{l c r rr rr rr}
\toprule
\multirow{2}{*}{\textbf{Group}} 
& \multirow{2}{*}{\textbf{Lang}} 
& \multirow{2}{*}{\textbf{Baseline}} 
& \multicolumn{2}{c}{$\tau=0.75$} 
& \multicolumn{2}{c}{$\tau=0.85$} 
& \multicolumn{2}{c}{$\tau=0.95$} \\
\cmidrule(lr){4-5} \cmidrule(lr){6-7} \cmidrule(lr){8-9}
& & 
& \textbf{F\textsubscript{1}} & $\boldsymbol{\Delta}$ 
& \textbf{F\textsubscript{1}} & $\boldsymbol{\Delta}$ 
& \textbf{F\textsubscript{1}} & $\boldsymbol{\Delta}$ \\
\midrule

aar & aar & $97.10$ & $\mathbf{98.24}$ & $\mathbf{+1.14}$ & $\mathbf{98.24}$ & $\mathbf{+1.14}$ & $\mathbf{98.24}$ & $\mathbf{+1.14}$ \\
\hline

bsq & bas & $94.23$ & $\mathbf{95.36}$ & $\mathbf{+1.13}$ & $\mathbf{95.40}$ & $\mathbf{+1.17}$ & $\mathbf{95.46}$ & $\mathbf{+1.23}$ \\
\hline

\multirow{5}{*}{ara} 
& aeb & $94.50$ & $94.68$ & $+0.18$ & $\mathbf{95.13}$ & $\mathbf{+0.63}$ & $95.00$ & $+0.50$ \\
& ara & $92.55$ & $92.84$ & $+0.29$ & $92.99$ & $+0.44$ & $\mathbf{93.12}$ & $\mathbf{+0.57}$ \\
& arq & $98.89$ & $99.21$ & $+0.32$ & $99.53$ & $+0.64$ & $\mathbf{99.68}$ & $\mathbf{+0.79}$ \\
& ary & $96.05$ & $96.16$ & $+0.11$ & $\mathbf{96.35}$ & $\mathbf{+0.31}$ & $96.17$ & $+0.13$ \\
& arz & $93.46$ & $95.70$ & $+2.24$ & $\mathbf{96.23}$ & $\mathbf{+2.77}$ & $98.91$ & $+5.45$ \\
\hline
\multirow{4}{*}{kin} 
& kin & $92.90$ & $93.17$ & $+0.27$ & $93.17$ & $+0.27$ & $\mathbf{93.57}$ & $\mathbf{+0.66}$ \\
& run & $95.34$ & $95.37$ & $+0.03$ & $95.37$ & $+0.03$ & $\mathbf{96.05}$ & $\mathbf{+0.71}$ \\
& swh & $99.80$ & $\mathbf{99.90}$ & $\mathbf{+0.10}$ & $\mathbf{99.90}$ & $\mathbf{+0.10}$ & $\mathbf{99.90}$ & $\mathbf{+0.10}$ \\
& nnb & $98.58$ & $\mathbf{98.72}$ & $\mathbf{+0.14}$ & $\mathbf{98.72}$ & $\mathbf{+0.14}$ & $\mathbf{98.72}$ & $\mathbf{+0.14}$ \\ 
\hline
 \multirow{3}{*}{lns} & tik & $99.60$ & $\mathbf{99.70}$ & $\mathbf{+0.10}$ & $\mathbf{99.70}$ & $\mathbf{+0.10}$ & $\mathbf{99.70}$ & $\mathbf{+0.10}$ \\
 & pfe & $96.83$ & $\mathbf{96.96}$ & $\mathbf{+0.13}$ & $\mathbf{96.96}$ & $\mathbf{+0.13}$ & $\mathbf{96.96}$ & $\mathbf{+0.13}$ \\
 & agq & $95.16$ & $\mathbf{95.29}$ & $\mathbf{+0.14}$ & $\mathbf{95.29}$ & $\mathbf{+0.14}$ & $\mathbf{95.29}$ & $\mathbf{+0.14}$ \\
\hline
 \multirow{3}{*}{blo} & kbp & $98.51$ & $\mathbf{98.64}$ & $\mathbf{+0.13}$ & $\mathbf{98.64}$ & $\mathbf{+0.13}$ & $\mathbf{98.64}$ & $\mathbf{+0.13}$ \\
 & dop & $99.17$ & $\mathbf{99.31}$ & $\mathbf{+0.14}$ & $\mathbf{99.31}$ & $\mathbf{+0.14}$ & $\mathbf{99.31}$ & $\mathbf{+0.14}$ \\
 & bib & $97.41$ & $\mathbf{97.54}$ & $\mathbf{+0.13}$ & $\mathbf{97.54}$ & $\mathbf{+0.13}$ & $\mathbf{99.35}$ & $\mathbf{+1.94}$ \\
\hline
\multirow{7}{*}{maf} 
& maf & $95.37$ & $\mathbf{99.50}$ & $\mathbf{+4.13}$ & $\mathbf{99.50}$ & $\mathbf{+4.13}$ & $\mathbf{99.50}$ & $\mathbf{+4.13}$ \\
& mcn & $96.05$ & $98.17$ & $+2.12$ & $98.37$ & $+2.32$ & $\mathbf{99.18}$ & $\mathbf{+3.13}$ \\
& meq & $95.26$ & $\mathbf{99.86}$ & $\mathbf{+4.60}$ & $\mathbf{99.86}$ & $\mathbf{+4.60}$ & $\mathbf{99.86}$ & $\mathbf{+4.60}$ \\
& mfz & $95.74$ & $97.79$ & $+2.05$ & $97.89$ & $+2.15$ & $\mathbf{98.84}$ & $\mathbf{+3.10}$ \\
& mpe & $96.19$ & $98.56$ & $+2.37$ & $98.76$ & $+2.57$ & $\mathbf{98.96}$ & $\mathbf{+2.77}$ \\
& mug & $96.60$ & $98.90$ & $+2.30$ & $99.30$ & $+2.70$ & $\mathbf{99.70}$ & $\mathbf{+3.10}$ \\
& myx & $94.96$ & $\mathbf{95.86}$ & $\mathbf{+0.90}$ & $\mathbf{95.86}$ & $\mathbf{+0.90}$ & $\mathbf{95.86}$ & $\mathbf{+0.90}$ \\
 \hline
din & din & $70.51$ & $\mathbf{73.75}$ & $\mathbf{+3.24}$ & $72.96$ & $+2.44$ & $\mathbf{73.75}$ & $\mathbf{+3.24}$ \\
\hline
ful & fub & $92.28$ & $95.62$ & $+3.34$ & $96.33$ & $+4.05$ & $\mathbf{96.59}$ & $\mathbf{+4.31}$ \\
\hline
sot & sot & $92.83$ & $\mathbf{94.08}$ & $\mathbf{+1.25}$ & $\mathbf{94.08}$ & $\mathbf{+1.25}$ & $\mathbf{94.08}$ & $\mathbf{+0.25}$ \\
\hline
ibo & ibo & $98.58$ & $\mathbf{99.53}$ & $\mathbf{+0.95}$ & $\mathbf{99.53}$ & $\mathbf{+0.95}$ & $\mathbf{99.53}$ & $\mathbf{+0.95}$ \\
\hline
wol & wol & $88.13$ & $\mathbf{90.35}$ & $\mathbf{+2.22}$ & $\mathbf{90.35}$ & $\mathbf{+2.22}$ & $\mathbf{90.35}$ & $\mathbf{+2.22}$ \\
\hline
swa & swc & $94.21$ & $94.55$ & $+0.34$ & $94.28$ & $+0.07$ & $\mathbf{94.77}$ & $\mathbf{+0.55}$ \\
\midrule
\textbf{Average} & & -- & -- & $\mathbf{+1.22}$ & -- & $\mathbf{+1.29}$ & -- & $\mathbf{+1.57}$ \\
\bottomrule
\end{tabular}
}
\caption{Hierarchical classification results using \ourmirror~embeddings across available confidence thresholds ($\tau=0.75, 0.85, 0.95$). Baseline F\textsubscript{1} shows base Serengeti performance; $\Delta$ columns show improvement over baseline. Bold indicates best performance per language.}
\label{tab:full_hier}
\end{table*}

%% file: tables/tab_top_bts.tex
\begin{table}[t]
\centering\small
\begin{tabular}{llrrc}
\toprule
Source & Target & BTS$^{F1}$ & BTS$^{\mathcal{L}}$ & Cousin \\
\midrule
\textit{bam} & \textit{dyu} & +0.22 & $-$0.08 & \checkmark \\
\textit{fub} & \textit{fuv} & +0.16 & $-$0.11 & \checkmark \\
\textit{nnb} & \textit{fuv} & +0.14 & $-$0.09 &  \\
\textit{dyu} & \textit{bam} & +0.14 & $-$0.17 & \checkmark \\
\textit{tsn} & \textit{fuv} & +0.13 & $-$0.08 &  \\
\textit{xho} & \textit{fuv} & +0.13 & $-$0.08 &  \\
\textit{fuh} & \textit{fuq} & +0.12 & $-$0.02 &  \\
\textit{afr} & \textit{fuv} & +0.12 & +0.01 &  \\
\textit{fub} & \textit{fuq} & +0.12 & +0.03 &  \\
\textit{tso} & \textit{fuv} & +0.12 & $-$0.12 &  \\
\bottomrule
\end{tabular}
\caption{Largest F\textsubscript{1} transfer events. Confusable cousins improve F\textsubscript{1} while degrading calibration (negative BTS$^{\mathcal{L}}$).}
\label{tab:top_bts}
\end{table}

%% file: tables/tab_script_pairs.tex
\begin{table}[t]
\centering\small
\begin{tabular}{lrrr}
\toprule
Script pair & $n$ & $\overline{|\mathrm{BTS}^{F1}|}$ & $\overline{|\mathrm{BTS}^{\mathcal{L}}|}$ \\
\midrule
\textbf{Latn+Latn} & 506 & 0.015 & 0.012 \\
Arab+Latn & 184 & 0.010 & 0.011 \\
Ethi+Latn & 138 & 0.008 & 0.008 \\
Arab+Ethi & 24 & 0.007 & 0.010 \\
\textbf{Arab+Arab} & 12 & 0.023 & 0.066 \\
\textbf{Ethi+Ethi} & 6 & 0.001 & 0.006 \\
\bottomrule
\end{tabular}
\caption{Mean transfer magnitude by script pair. Within-script pairs (bold) dominate; the within-Arabic block carries the largest calibration effects.}
\label{tab:script_pairs}
\end{table}

%% file: tables/supported_lang.tex

\begin{table*}[t]
\centering
\scriptsize
\setlength{\tabcolsep}{3.0pt}
\renewcommand{\arraystretch}{0.88}
\resizebox{\textwidth}{!}{%
\begin{tabular}{@{}>{\bfseries}l l r@{\hspace{1.4em}}>{\bfseries}l l r@{\hspace{1.4em}}>{\bfseries}l l r@{}}
\toprule
ISO-3 & Language & Sentences & ISO-3 & Language & Sentences & ISO-3 & Language & Sentences \\
\midrule
aar & Afar & 2,930 & aba & Abé & 6,902 & abi & Abidji & 10,218 \\
abn & Abua & 8,431 & acd & Gikyode & 7,102 & ach & Acholi & 8,626 \\
ada & Dangme & 11,002 & ade & Adele & 14,293 & adh & Jopadhola & 7,569 \\
adj & Adioukrou & 7,121 & aeb & Arabic, Tunisian & 12,985 & afr & Afrikaans & 45,990 \\
agq & Aghem & 2,942 & aha & Ahanta & 6,773 & ajg & Aja & 7,714 \\
aka & Akan & 2,948 & akp & Siwu & 7,118 & ald & Alladian & 13,794 \\
alz & Alur & 10,003 & amf & Hamer-Banne & 14,270 & amh & Amharic & 35,740 \\
ann & Obolo & 7,310 & anu & Anuak & 2,917 & anv & Denya & 7,126 \\
any & Anyin & 13,994 & ara & Arabic & 2,938 & arq & Arabic, Algerian & 3,693 \\
ary & Arabic, Moroccan & 24,745 & arz & Arabic, Egyptian & 25,958 & asa & Asu & 2,939 \\
asg & Cishingini & 6,884 & atg & Ivbie North-Okpela-Arhe & 7,135 & ati & Attié & 6,212 \\
avn & Avatime & 6,638 & avu & Avokaya & 7,080 & azo & Awing & 2,627 \\
bam & Bamanankan & 11,854 & bas & Basaa & 8,895 & bav & Vengo & 6,800 \\
bba & Baatonum & 9,452 & bbj & Ghomálá’ & 6,605 & bbk & Babanki & 5,191 \\
bbo & Konabéré & 10,168 & bci & Baoulé & 12,880 & bcn & Bali & 2,942 \\
bcw & Bana & 7,119 & bcy & Bacama & 336 & bdh & Baka & 6,750 \\
bds & Burunge & 2,940 & bem & Bemba & 41,323 & beq & Beembe & 6,638 \\
bex & Jur Modo & 7,389 & bez & Bena & 2,946 & bfa & Bari & 2,936 \\
bfd & Bafut & 7,129 & bfo & Birifor, Malba & 6,824 & bib & Bisa & 7,133 \\
bim & Bimoba & 8,827 & bin & Edo & 9,811 & biv & Birifor, Southern & 7,124 \\
bjv & Bedjond & 7,569 & bkv & Bekwarra & 14,291 & bky & Bokyi & 2,929 \\
blh & Kuwaa & 10,202 & bmo & Chrambo & 2,940 & bmq & Bomu & 14,293 \\
bmv & Bum & 6,633 & bom & Berom & 6,676 & bov & Tuwuli & 7,117 \\
box & Buamu & 7,103 & bqc & Boko & 8,703 & bqj & Bandial & 6,635 \\
bqp & Bisã & 14,290 & bsc & Oniyan & 6,636 & bsp & Baga Sitemu & 6,332 \\
bsq & Bassa & 4,803 & bss & Akoose & 7,119 & bst & Basketo & 1,435 \\
btt & Bete-Bendi & 14,286 & bud & Ntcham & 6,825 & bum & Bulu & 10,172 \\
bun & Sherbro & 333 & bus & Bokobaru & 7,213 & buy & Bullom So & 668 \\
bwq & Bobo Madaré, Southern & 14,292 & bwr & Bura-Pabir & 2,929 & bwu & Buli & 7,114 \\
bxk & Bukusu & 2,904 & byf & Bete & 2,681 & byv & Medumba & 3,902 \\
bza & Bandi & 2,942 & bzw & Basa & 2,939 & cce & Chopi & 9,805 \\
cgg & Chiga & 5,262 & chw & Chuwabu & 9,742 & cjk & Chokwe & 21,845 \\
cko & Anufo & 6,786 & cme & Cerma & 7,105 & cop & Coptic & 9,302 \\
cou & Wamey & 6,042 & cri & Sãotomense & 2,697 & crs & Seychelles French Creole & 13,500 \\
csk & Jola-Kasa & 7,127 & cwe & Kwere & 7,120 & cwt & Kuwaataay & 14,288 \\
daa & Dangaléat & 7,104 & daf & Dan & 13,673 & dag & Dagbani & 9,656 \\
dav & Dawida & 2,944 & dbq & Daba & 10,081 & ddn & Dendi & 2,238 \\
dga & Dagaare, Southern & 6,991 & dgd & Dagaari Dioula & 2,942 & dgi & Dagara, Northern & 6,829 \\
dhm & Dhimba & 6,634 & dib & Dinka, South Central & 1,175 & did & Didinga & 5,898 \\
dig & Chidigo & 6,828 & dik & Dinka, Southwestern & 17,349 & din & Dinka & 340 \\
dip & Dinka, Northeastern & 6,051 & diu & Gciriku & 906 & dje & Zarma & 2,749 \\
dks & Dinka, Southeastern & 6,965 & dnj & Dan & 7,216 & dop & Lukpa & 15,230 \\
dos & Dogosé & 10,232 & dow & Doyayo & 6,643 & dsh & Daasanach & 6,019 \\
dts & Dogon, Toro So & 14,276 & dua & Duala & 8,520 & dug & Chiduruma & 6,824 \\
dur & Dii & 13,791 & dwr & Dawro & 7,582 & dyi & Sénoufo, Djimini & 7,105 \\
dyo & Jola-Fonyi & 3,800 & dyu & Jula & 20,860 & ebr & Tchaman & 2,939 \\
ebu & Kiembu & 2,915 & efi & Efik & 16,007 & ego & Eggon & 2,942 \\
eka & Ekajuk & 7,120 & eko & Koti & 4,636 & enb & Markweeta & 13,700 \\
\bottomrule
\end{tabular}%
}
\caption{Languages covered in \ourdata -- Part I}
\label{tab:train-balance-10m-languages-i}
\end{table*}
\clearpage

\begin{table*}[t]
\centering
\scriptsize
\setlength{\tabcolsep}{3.0pt}
\renewcommand{\arraystretch}{0.88}
\resizebox{\textwidth}{!}{%
\begin{tabular}{@{}>{\bfseries}l l r@{\hspace{1.4em}}>{\bfseries}l l r@{\hspace{1.4em}}>{\bfseries}l l r@{}}
\toprule
ISO-3 & Language & Sentences & ISO-3 & Language & Sentences & ISO-3 & Language & Sentences \\
\midrule
eto & Eton & 2,286 & etu & Ejagham & 6,635 & etx & Iten & 2,936 \\
ewe & Éwé & 42,255 & ewo & Ewondo & 6,829 & fak & Fang & 2,854 \\
fal & Fali, South & 14,292 & fan & Fang & 14,513 & ffm & Fulfulde, Maasina & 7,508 \\
fia & Nobiin & 345 & fip & Fipa & 2,944 & flr & Fuliiru & 2,924 \\
fon & Fon & 22,482 & fub & Fulfulde, Adamawa & 8,041 & fue & Fulfulde, Borgu & 6,615 \\
fuf & Pular & 7,355 & fuh & Fulfulde, Western Niger & 7,154 & ful & Fulah & 2,513 \\
fuq & Fulfulde, Central-Eastern Niger & 6,737 & fuv & Fulfulde, Nigerian & 16,142 & gaa & Ga & 16,520 \\
gax & Oromo, Borana-Arsi-Guji & 2,852 & gaz & Oromo, West Central & 37,860 & gbo & Grebo, Northern & 5,684 \\
gbr & Gbagyi & 6,617 & gde & Gude & 7,118 & gej & Gen & 10,257 \\
gid & Gidar & 6,628 & giz & Giziga & 7,947 & gjn & Gonja & 7,833 \\
gkn & Gokana & 9,191 & gkp & Kpelle, Guinea & 2,898 & gmv & Gamo & 7,987 \\
gna & Kaansa & 6,614 & gnd & Zulgo-Gemzek & 6,815 & gng & Ngangam & 7,106 \\
goa & Guro & 1,929 & gof & Gofa & 7,470 & gog & Gogo & 6,869 \\
gol & Gola & 2,941 & gqr & Gor & 7,129 & gso & Gbaya, Southwest & 7,168 \\
gud & Dida, Yocoboué & 6,629 & guk & Gumuz & 14,292 & gur & Farefare & 9,570 \\
guw & Gun & 14,422 & gux & Gourmanchéma & 7,617 & guz & Ekegusii & 6,546 \\
gvl & Gulay & 6,918 & gwr & Gwere & 6,046 & gya & Gbaya, Northwest & 8,038 \\
hae & Oromo, Eastern & 14,293 & hag & Hanga & 4,840 & har & Harari & 114 \\
hau & Hausa & 32,959 & hav & Havu & 11,013 & hay & Haya & 6,848 \\
hbb & Nya Huba & 2,939 & heh & Hehe & 7,169 & her & Herero & 10,708 \\
hgm & Hai|ǁom & 2,888 & hig & Kamwe & 3,696 & hna & Mina & 2,937 \\
ibb & Ibibio & 2,945 & ibo & Igbo & 35,306 & idu & Idoma & 8,803 \\
ife & Ifè & 14,528 & igb & Ebira & 2,942 & ige & Igede & 8,879 \\
igl & Igala & 2,947 & ijn & Kalabari & 2,932 & ikk & Ika & 8,221 \\
ikw & Ikwere & 7,128 & ilb & Ila & 13,584 & iqw & Ikwo & 6,631 \\
iri & Rigwe & 7,130 & irk & Iraqw & 14,292 & ish & Esan & 9,493 \\
iso & Isoko & 14,172 & iyx & Yaka & 853 & izr & Izere & 7,132 \\
izz & Izii & 8,063 & jbu & Jukun Takum & 14,292 & jgo & Ngomba & 2,939 \\
jib & Jibu & 2,933 & jit & Jita & 2,939 & jmc & Machame & 6,752 \\
kab & Kabyle & 36,057 & kam & Kamba & 24,505 & kao & Xaasongaxango & 14,292 \\
kbn & Kare & 2,939 & kbo & Keliko & 5,192 & kbp & Kabiyè & 26,519 \\
kbr & Kafa & 14,182 & kby & Kanuri, Manga & 2,502 & kcg & Tyap & 6,639 \\
kck & Kalanga & 7,593 & kdc & Kutu & 7,122 & kde & Makonde & 7,631 \\
kdh & Tem & 4,320 & kdi & Kumam & 7,020 & kdj & Ng’akarimojong & 6,717 \\
kdl & Tsikimba & 7,119 & kdn & Kunda & 2,924 & kea & Kabuverdianu & 21,334 \\
ken & Kenyang & 7,120 & keo & Kakwa & 2,363 & ker & Kera & 10,211 \\
kez & Kukele & 14,291 & khq & Songhay, Koyra Chiini & 10,230 & khy & Kele & 6,639 \\
kia & Kim & 8,214 & kik & Gikuyu & 28,959 & kin & Kinyarwanda & 53,340 \\
kiz & Kisi & 2,885 & kki & Kagulu & 7,132 & kkj & Kako & 7,127 \\
kln & Kalenjin & 2,871 & klu & Klao & 2,932 & kma & Konni & 6,042 \\
kmb & Kimbundu & 26,546 & kmy & Koma & 5,307 & knc & Kanuri, Yerwa & 10,991 \\
knf & Mankanya & 7,194 & kng & Koongo & 7,442 & knk & Kuranko & 6,818 \\
kno & Kono & 6,805 & kny & Kanyok & 5,261 & kon & Kongo & 2,948 \\
koo & Konzo & 11,112 & koq & Kota & 390 & kpz & Kupsapiiny & 3,741 \\
kqn & Kaonde & 13,753 & kqo & Krahn, Eastern & 13,800 & kqp & Kimré & 7,125 \\
kqs & Kissi, Northern & 6,629 & kqy & Koorete & 6,796 & kri & Krio & 10,861 \\
krs & Gbaya & 2,947 & krw & Krahn, Western & 2,932 & krx & Karon & 1,201 \\
ksb & Shambala & 6,627 & ksf & Bafia & 6,655 & ksp & Kabba & 6,383 \\
\bottomrule
\end{tabular}%
}
\caption{Languages covered in \ourdata -- Part II}
\label{tab:train-balance-10m-languages-ii}
\end{table*}
\clearpage

\begin{table*}[t]
\centering
\scriptsize
\setlength{\tabcolsep}{3.0pt}
\renewcommand{\arraystretch}{0.88}
\resizebox{\textwidth}{!}{%
\begin{tabular}{@{}>{\bfseries}l l r@{\hspace{1.4em}}>{\bfseries}l l r@{\hspace{1.4em}}>{\bfseries}l l r@{}}
\toprule
ISO-3 & Language & Sentences & ISO-3 & Language & Sentences & ISO-3 & Language & Sentences \\
\midrule
kss & Kisi, Southern & 7,796 & ktb & Kambaata & 14,253 & ktj & Krumen, Plapo & 7,098 \\
ktu & Kituba & 34,578 & kua & Oshiwambo & 13,681 & kub & Kutep & 7,127 \\
kuj & Kuria & 6,616 & kus & Kusaal & 6,834 & kvj & Psikye & 6,573 \\
kwn & Kwangali & 12,072 & kwy & Kikongo & 20,333 & kxc & Konso & 14,289 \\
kyf & Kouya & 6,851 & kyq & Kenga & 7,512 & kzn & Kokola & 2,789 \\
kzr & Karang & 2,695 & lai & Lambya & 6,618 & laj & Lango & 7,356 \\
lam & Lamba & 8,235 & lap & Laka & 4,931 & las & Lama & 10,206 \\
ldi & Laari & 15,818 & lea & Lega-Shabunda & 4,750 & led & Lendu & 5,573 \\
lee & Lyélé & 7,109 & lef & Lelemi & 6,826 & leh & Lenje & 15,944 \\
lem & Nomaande & 7,128 & lgg & Lugbara & 8,842 & lgm & Lega-Mwenga & 6,631 \\
lia & Limba, West-Central & 6,832 & lik & Lika & 2,929 & lin & Lingala & 43,654 \\
lip & Sekpele & 7,126 & llb & Lolo & 14,888 & lln & Lele & 12,030 \\
lmd & Lumun & 2,656 & lmp & Limbum & 6,628 & lnl & Banda, South Central & 2,924 \\
lob & Lobi & 14,293 & log & Logo & 4,561 & lok & Loko & 10,225 \\
lol & Mongo-Nkundu & 13,799 & lom & Loma & 6,396 & loq & Lobala & 6,308 \\
lot & Otuho & 2,939 & loz & Lozi & 15,985 & lro & Laro & 2,939 \\
lsm & Saamya-Gwe & 6,815 & lth & Thur & 2,895 & lto & Olutsotso & 2,885 \\
lua & Luba-Kasai & 39,225 & lub & Luba-Katanga & 14,269 & luc & Aringa & 6,501 \\
lue & Luvale & 14,312 & lug & Ganda & 31,230 & lun & Lunda & 13,024 \\
luo & Dholuo & 32,868 & lwg & Oluwanga & 3,411 & lwo & Luwo & 7,107 \\
maf & Mafa & 6,642 & mas & Maasai & 8,400 & maw & Mampruli & 7,113 \\
mbu & Mbula-Bwazza & 2,945 & mck & Mbunda & 10,430 & mcn & Masana & 7,797 \\
mcp & Makaa & 7,356 & mcu & Mambila, Cameroon & 6,825 & mda & Mada & 7,128 \\
mdm & Mayogo & 2,932 & mdy & Male & 8,977 & men & Mende & 6,891 \\
meq & Merey & 7,120 & mer & Kimîîru & 4,678 & mev & Maan & 2,415 \\
mfe & Morisyen & 13,701 & mfg & Mogofin & 4,216 & mfh & Matal & 6,833 \\
mfi & Wandala & 7,118 & mfk & Mofu, North & 7,130 & mfq & Moba & 6,677 \\
mfz & Mabaan & 4,007 & mgc & Morokodo & 5,449 & mgh & Makhuwa-Meetto & 9,323 \\
mgo & Meta’ & 6,638 & mgq & Malila & 2,939 & mgr & Mambwe-Lungu & 10,507 \\
mgw & Matumbi & 1,636 & mhi & Ma’di & 3,722 & mhw & Mbukushu & 45 \\
mif & Mofu-Gudur & 7,103 & mkl & Mokole & 7,119 & mlg & Malagasy & 2,481 \\
mlr & Vame & 887 & mmy & Migaama & 2,710 & mnf & Mundani & 6,821 \\
mnk & Mandinka & 6,832 & mny & Manyawa & 15,521 & moa & Mwan & 7,127 \\
mor & Moro & 3,695 & mos & Moore & 30,772 & moy & Shekkacho & 2,945 \\
moz & Mukulu & 2,928 & mpe & Majang & 2,932 & mpg & Marba & 7,119 \\
mqb & Mbuko & 7,122 & msc & Maninka, Sankaran & 5,067 & mse & Musey & 13,796 \\
mua & Mundang & 13,796 & mug & Musgu & 15,112 & muh & Mündü & 14,539 \\
mur & Murle & 6,814 & muy & Muyang & 6,933 & mwe & Mwera & 2,946 \\
mwm & Sar & 7,955 & mwn & Nyamwanga & 8,332 & mws & Mwimbi-Muthambi & 877 \\
myb & Mbay & 7,124 & myk & Sénoufo, Mamara & 7,155 & myx & Masaaba & 9,251 \\
mzk & Mambila, Nigeria & 14,293 & mzm & Mumuye & 7,139 & mzw & Deg & 7,089 \\
naq & Khoekhoe & 9,663 & naw & Nawuri & 7,133 & nba & Nyemba & 10,339 \\
nbl & Ndebele & 13,722 & ncu & Chumburung & 7,081 & ndc & Ndau & 11,260 \\
nde & Ndebele & 11,979 & ndh & Ndali & 2,455 & ndi & Samba Leko & 13,793 \\
ndj & Ndamba & 7,135 & ndo & Ndonga & 13,803 & ndp & Kebu & 14,293 \\
ndv & Ndut & 2,728 & ndy & Luto & 10,245 & ndz & Ndogo & 7,130 \\
neb & Toura & 14,284 & nfr & Nafaanra & 14,290 & ngb & Ngbandi, Northern & 4,833 \\
ngc & Ngombe & 6,639 & ngl & Lomwe & 10,462 & ngn & Ngwo & 2,899 \\
\bottomrule
\end{tabular}%
}
\caption{Languages covered in \ourdata -- Part III}
\label{tab:train-balance-10m-languages-iii}
\end{table*}
\clearpage

\begin{table*}[t]
\centering
\scriptsize
\setlength{\tabcolsep}{3.0pt}
\renewcommand{\arraystretch}{0.88}
\resizebox{\textwidth}{!}{%
\begin{tabular}{@{}>{\bfseries}l l r@{\hspace{1.4em}}>{\bfseries}l l r@{\hspace{1.4em}}>{\bfseries}l l r@{}}
\toprule
ISO-3 & Language & Sentences & ISO-3 & Language & Sentences & ISO-3 & Language & Sentences \\
\midrule
ngp & Ngulu & 7,137 & nhr & Naro & 7,113 & nhu & Noone & 7,111 \\
nih & Nyiha, Tanzania & 2,939 & nim & Nilamba & 7,137 & nin & Ninzo & 7,253 \\
niq & Nandi & 7,786 & niy & Ngiti & 6,638 & nka & Nkoya & 2,842 \\
nko & Nkonya & 7,107 & nla & Ngombale & 2,341 & nmz & Nawdm & 14,289 \\
nnb & Nande & 18,686 & nnh & Ngiemboon & 6,324 & nnq & Ngindo & 7,133 \\
nnw & Nuni, Southern & 14,526 & nqo & N’Ko & 12,823 & nse & Nsenga & 9,637 \\
nso & Sotho, Northern & 46,564 & ntr & Delo & 7,129 & nuj & Nyole & 7,658 \\
nus & Nuer & 11,861 & nwb & Nyabwa & 6,776 & nxd & Ngando & 6,644 \\
nya & Chichewa & 51,207 & nyb & Nyagbo & 2,940 & nyd & Olunyole & 2,942 \\
nyf & Kigiryama & 6,848 & nyk & Nyaneka & 11,885 & nym & Nyamwezi & 2,934 \\
nyn & Nyankore & 11,475 & nyo & Nyoro & 6,841 & nyu & Nyungwe & 10,624 \\
nyy & Nyakyusa-Ngonde & 9,740 & nza & Mbembe, Tigon & 5,188 & nzi & Nzema & 12,609 \\
odu & Odual & 2,938 & ogo & Khana & 8,464 & oke & Okpe & 8,523 \\
okr & Kirike & 2,941 & oku & Oku & 6,048 & old & Mochi & 13,340 \\
orm & Oromo & 2,792 & ozm & Koonzime & 6,799 & pbi & Parkwa & 14,279 \\
pcm & Pidgin, Nigerian & 13,317 & pem & Phende & 6,198 & pfe & Pere & 14,395 \\
phm & Phimbi & 15,399 & pkb & Kipfokomu & 6,828 & pko & Pökoot & 2,783 \\
plt & Malagasy, Merina & 27,852 & pny & Pinyin & 13,238 & pov & Guinea-Bissau Creole & 7,251 \\
poy & Pogolo & 6,644 & rag & Lulogooli & 2,900 & rcf & Réunion French Creole & 13,722 \\
rel & Rendille & 6,056 & rif & Tarifit & 2,941 & rim & Nyaturu & 7,137 \\
rnd & Ruund & 7,481 & rng & Ronga & 8,976 & rub & Gungu & 6,636 \\
ruf & Luguru & 14,533 & run & Rundi & 44,213 & rwk & Rwa & 2,633 \\
sag & Sango & 36,544 & saq & Samburu & 2,946 & sba & Ngambay & 8,083 \\
sbd & Samo, Southern & 6,654 & sbp & Sangu & 2,939 & sbs & Kuhane & 2,540 \\
sby & Soli & 2,407 & sef & Sénoufo, Cebaara & 2,935 & seh & Sena & 19,145 \\
ses & Songhay, Koyraboro Senni & 7,372 & sev & Sénoufo, Nyarafolo & 2,930 & sfw & Esahie & 6,359 \\
sgc & Kipsigis & 11,612 & sgw & Sebat Bet Gurage & 6,828 & shi & Tachelhit & 8,390 \\
shj & Shatt & 766 & shk & Shilluk & 5,943 & shr & Shi & 14,113 \\
shu & Arabic, Chadian & 6,626 & sid & Sidaama & 9,955 & sig & Paasaal & 7,124 \\
sil & Sisaala, Tumulung & 6,827 & skg & Malagasy, Sakalava & 15,615 & sld & Sissala & 14,290 \\
sna & Shona & 45,067 & snf & Noon & 6,634 & sng & Sanga & 2,913 \\
snw & Selee & 6,863 & soe & Ohendo & 717 & som & Somali & 27,870 \\
sop & Songe & 11,004 & sor & Soumraye & 1,083 & sot & Sotho, Southern & 13,079 \\
soy & Miyobe & 7,118 & spp & Sénoufo, Supyire & 7,114 & spy & Sabaot & 14,640 \\
srr & Serer-Sine & 3,905 & ssw & Swati & 27,224 & suk & Sukuma & 7,344 \\
sur & Mwaghavul & 14,293 & sus & Susu & 8,885 & swa & Swahili & 2,918 \\
swb & Comorian, Maore & 237 & swc & Swahili, Congo & 28,449 & swh & Swahili & 37,548 \\
swk & Sena, Malawi & 5,962 & sxb & Suba & 7,121 & tap & Taabwa & 10,437 \\
taq & Tamasheq & 15,981 & tbz & Ditammari & 5,589 & tcc & Datooga & 7,128 \\
tcd & Tafi & 2,947 & tdx & Malagasy, Tandroy-Mahafaly & 8,366 & ted & Krumen, Tepo & 6,861 \\
tem & Themne & 6,840 & teo & Ateso & 8,489 & tex & Tennet & 2,942 \\
tgw & Sénoufo, Tagwana & 2,934 & thk & Kitharaka & 6,820 & thv & Tamahaq, Tahaggart & 1,946 \\
tig & Tigré & 3,105 & tik & Tikar & 14,289 & tir & Tigrigna & 36,754 \\
tiv & Tiv & 13,164 & tke & Takwane & 6,888 & tlj & Talinga-Bwisi & 7,117 \\
tll & Tetela & 13,894 & tmc & Tumak & 10,258 & tnr & Ménik & 9,951 \\
tod & Toma & 7,199 & tog & Tonga & 11,237 & toh & Tonga & 9,554 \\
toi & Tonga & 15,080 & tpm & Tampulma & 8,553 & tsc & Tswa & 12,065 \\
tsn & Setswana & 35,384 & tso & Tsonga & 38,043 & tsw & Tsishingini & 7,125 \\
\bottomrule
\end{tabular}%
}
\caption{Languages covered in \ourdata -- Part IV}
\label{tab:train-balance-10m-languages-iv}
\end{table*}
\clearpage

\begin{table*}[t]
\centering
\scriptsize
\setlength{\tabcolsep}{3.0pt}
\renewcommand{\arraystretch}{0.88}
\resizebox{\textwidth}{!}{%
\begin{tabular}{@{}>{\bfseries}l l r@{\hspace{1.4em}}>{\bfseries}l l r@{\hspace{1.4em}}>{\bfseries}l l r@{}}
\toprule
ISO-3 & Language & Sentences & ISO-3 & Language & Sentences & ISO-3 & Language & Sentences \\
\midrule
ttj & Tooro & 9,664 & ttq & Tamajaq, Tawallammat & 6,624 & ttr & Tera & 2,946 \\
tui & Tupuri & 7,698 & tul & Tula & 5,200 & tum & Tumbuka & 34,918 \\
tuv & Turkana & 2,383 & tvu & Tunen & 2,942 & twi & Twi & 43,976 \\
twx & Tewe & 14,947 & tzm & Tamazight, Central Atlas & 2,472 & udu & Uduk & 4,024 \\
umb & Umbundu & 32,657 & urh & Urhobo & 10,605 & uth & ut-Hun & 5,189 \\
vag & Vagla & 7,010 & vai & Vai & 2,940 & ven & Venda & 17,236 \\
vid & Vidunda & 7,117 & vif & Vili & 2,942 & vmk & Makhuwa-Shirima & 4,231 \\
vmw & Makhuwa & 11,968 & vun & Vunjo & 7,131 & vut & Vute & 7,120 \\
wal & Wolaytta & 12,310 & wbi & Vwanji & 2,937 & wec & Wè Western & 2,691 \\
wes & Pidgin, Cameroon & 8,959 & wib & Toussian, Southern & 6,158 & wlx & Wali & 14,628 \\
wmw & Mwani & 7,125 & wob & Wè Northern & 14,529 & wol & Wolof & 17,637 \\
won & Wongo & 512 & wwa & Waama & 3,694 & xan & Xamtanga & 1,474 \\
xed & Hdi & 7,103 & xho & Xhosa & 37,847 & xmv & Malagasy, Antankarana & 15,974 \\
xnz & Mattokki & 2,939 & xog & Soga & 7,860 & xon & Konkomba & 7,511 \\
xpe & Kpelle, Liberia & 2,864 & xrb & Karaboro, Eastern & 6,933 & xsm & Kasem & 7,519 \\
xtc & Katcha-Kadugli-Miri & 2,946 & xuo & Kuo & 7,113 & yal & Yalunka & 7,086 \\
yam & Yamba & 7,124 & yao & Yao & 11,100 & yas & Nugunu & 2,168 \\
yat & Yambeta & 6,027 & yaz & Lokaa & 10,258 & yba & Yala & 2,744 \\
ybb & Yemba & 6,301 & yom & Kiyombe & 9,055 & yor & Yoruba & 48,191 \\
yre & Yaouré & 7,668 & zaj & Zaramo & 2,937 & zdj & Comorian, Ngazidja & 4,106 \\
zga & Kinga & 2,941 & zgh & Tamazight, Standard Moroccan & 5,817 & ziw & Zigula & 7,137 \\
zne & Zande & 11,952 & zul & Zulu & 37,885 &  &  &  \\
\bottomrule
\end{tabular}%
}
\caption{Languages covered in \ourdata -- Part V}
\label{tab:train-balance-10m-languages-v}
\end{table*}